\documentclass[sigconf]{acmart}
\usepackage{amsmath}
\usepackage[utf8]{inputenc} % allow utf-8 input
\usepackage[T1]{fontenc}    % use 8-bit T1 fonts
\usepackage{hyperref}       % hyperlinks
\usepackage{url}            % simple URL typesetting
\usepackage{booktabs}       % professional-quality tables
\usepackage{amsfonts}       % blackboard math symbols
\usepackage{nicefrac}       % compact symbols for 1/2, etc.
\usepackage{microtype}      % microtypography
\usepackage{xcolor}         % colors
\usepackage{graphicx}
\usepackage{wrapfig}
\usepackage{amsmath}
\usepackage{framed}
\usepackage{tcolorbox}      % 引入 tcolorbox 包
\tcbuselibrary{breakable}   % 启用 breakable 功能
\usepackage{xcolor}
\usepackage{mdframed}
\usepackage{lipsum}
\usepackage{subcaption}
\usepackage{graphicx}
\usepackage{enumitem} 
\usepackage{multirow}
\usepackage{xcolor}
\usepackage[table,xcdraw]{xcolor}
\usepackage{caption}
\usepackage{xurl}

\newcommand{\paratitle}[1]{\noindent\textbf{#1}}
\usepackage{makecell}
\usepackage{colortbl}
\usepackage{hhline}
\tcbuselibrary{breakable}
\AtBeginDocument{%
  }

\copyrightyear{2026}
\acmYear{2026}
\setcopyright{cc}
\setcctype{by}
\acmConference[WWW '26]{Proceedings of the ACM Web Conference 2026}{April 13--17, 2026}{Dubai, United Arab Emirates}
\acmBooktitle{Proceedings of the ACM Web Conference 2026 (WWW '26), April 13--17, 2026, Dubai, United Arab Emirates}
\acmPrice{}
\acmDOI{10.1145/XXX.XXX}
\acmISBN{979-8-4007-2307-0/2026/04}

\begin{document}

\title{Can LLMs Fool Graph Learning? Exploring Universal Adversarial Attacks on Text-Attributed Graphs}

\author{Zihui Chen}
\authornote{Both authors contributed equally to this research.}
\email{231270002@hdu.edu.cn}
\author{Yuling Wang}
\authornotemark[1]
\email{wangyl0612@hdu.edu.cn}
\affiliation{%
  \institution{School of Cyberspace, Hangzhou Dianzi University}
  \city{Hangzhou}
  \country{China}
}

\author{Pengfei Jiao}\authornote{Corresponding author.}
\affiliation{%
  \institution{  School of Cyberspace, Hangzhou Dianzi University }
  \city{Hangzhou}
  \country{China}}
\email{pjiao@hdu.edu.cn}

\author{Kai Wu}
\affiliation{%
  \institution{Hangzhou Dianzi University }
  \city{Hangzhou}
  \country{China}}
\email{wukai@hdu.edu.cn}

\author{Xiao Wang}
\affiliation{%
  \institution{Beihang University}
  \city{Beijing}
  \country{China}}
\email{xiao_wang@buaa.edu.cn}

\author{Xiang Ao}
\affiliation{%
  \institution{Institute of Computing Technology, Chinese Academy of Sciences}
  \city{Beijing}
  \country{China}}
\email{aoxiang@ict.ac.cn}

\author{Dalin Zhang}
\affiliation{%
  \institution{Hangzhou Dianzi University}
  \city{Hangzhou}
  \country{China}}
\email{dalinz@cs.aau.dk}

\renewcommand{\shortauthors}{Zihui Chen et al.}

\begin{abstract}

  Text-attributed graphs (TAGs) enhance graph learning by integrating rich textual semantics and topological context for each node. While boosting expressiveness, they also expose new vulnerabilities in graph learning through text-based adversarial surfaces. Recent advances leverage diverse backbones, such as graph neural networks (GNNs) and pre-trained language models (PLMs), to capture both structural and textual information in TAGs. This diversity raises a key question: \textit{How can we design universal adversarial attacks that generalize across architectures to assess the security of TAG models?} The challenge arises from the stark contrast in how different backbones—GNNs and PLMs—perceive and encode graph patterns, coupled with the fact that many PLMs are only accessible via APIs, limiting attacks to black-box settings. To address this, we propose \textsc{BadGraph}, a novel attack framework that deeply elicits large language models’ (LLMs) understanding of general graph knowledge to jointly perturb both node topology and textual semantics. Specifically, we design a target influencer retrieval module that leverages graph priors to construct cross-modally aligned attack shortcuts, thereby enabling efficient LLM-based perturbation reasoning. Experiments show that \textsc{BadGraph} achieves universal and effective attacks across GNN- and LLM-based reasoners, with up to a 76.3\% performance drop, while theoretical and empirical analyses confirm its stealthy yet interpretable nature.

\end{abstract}

% \begin{CCSXML}
% <ccs2012>
%   <concept>
%     <concept_id>10010147.10010257.10010258.10010260</concept_id>
%     <concept_desc>Computing methodologies~Machine learning</concept_desc>
%     <concept_significance>500</concept_significance>
%   </concept>
% <ccs2012>
% \end{CCSXML}
% \ccsdesc[500]{Computing methodologies~Machine learning}

\begin{CCSXML}
<ccs2012>
   <concept>
       <concept_id>cssClassifiers^500</concept_id>
       <concept_desc></concept_desc>
       <concept_significance>cssClassifiers^500</concept_significance>
       </concept>
   <concept>
       <concept_id>10010147.10010257.10010293</concept_id>
       <concept_desc>Computing methodologies~Machine learning approaches</concept_desc>
       <concept_significance>500</concept_significance>
       </concept>
 </ccs2012>
\end{CCSXML}
\ccsdesc[500]{Computing methodologies~Machine learning approaches}

\keywords{Graph Machine Learning, Text-Attributed Graph Attacks}

\maketitle
% \textbf{Statement of Relevance.} 
% This work presents an LLM-driven framework for attacking Web-related graphs, advancing the robustness and trustworthiness of graph algorithms, and is highly relevant to the \textit{Graph Algorithms and Modeling for the Web} track.
\section{ Introduction}

% Graphs, widely used in domains such as biological networks \cite{ma2023single}  and knowledge graphs \cite{san2024kg}, effectively model complex relationships among entities \cite{schlichtkrull2018modeling}. 
% Growing evidence suggests that graph learning models are highly vulnerable to adversarial attacks \cite{chen2022understandingimprovinggraphinjection}. 
% In many real-world scenarios, nodes within these graphs are enriched with descriptive textual attributes, resulting in text-attributed graphs (TAGs) \cite{yan2023comprehensive}. While this fusion of structural and semantic information enhances representational power, it also introduces new vulnerabilities through the textual modality, exposing novel adversarial attack surfaces \cite{lei2024intruding}. Consequently, developing tailored adversarial attacks  for TAGs is critical to ensuring the secure and trustworthy deployment of graph learning systems in safety-critical domains.

Graph representation learning has emerged as a powerful paradigm for modeling intricate relationships among entities in complex real-world systems, such as knowledge graphs \cite{san2024kg} and biological networks \cite{ma2023single}. Building on this, text-attributed graphs (TAGs) enhance each node with descriptive textual information, seamlessly integrating structural and semantic features \cite{yan2023comprehensive}. This fusion significantly boosts the expressive power of graph representations, enabling more detailed understanding and analysis. Consequently, research on TAGs has gained strong momentum across diverse fields, including graph neural networks (GNNs)~\cite{zhu2025graphclipenhancingtransferabilitygraph}, natural language processing~\cite{guo2025lightragsimplefastretrievalaugmented, zhang2021eatn}, and recommender systems~\cite{li2025g,zhang2021multi,wang2024can}.

% TAG modeling
Bridging topological and textual knowledge through effective alignment and fusion is essential for learning on TAGs~\cite{liu2025graph}. Broadly, existing efforts fall into two main paradigms:
(1) GNN as a reasoner. In this paradigm, node-associated texts are first encoded into feature using language models, and then propagated through the graph topology via GNN message-passing. 
The quality of text encoding is crucial. As a result, research has shifted from shallow methods like TF-IDF~\cite{kipf2016semi} to powerful large language models (LLMs) such as OFA~\cite{liu2023one} and TAPE~\cite{he2024harnessing}.
(2) LLM as a reasoner. This paradigm reformats graph data into a format that LLMs can understand, then either prompts frozen LLMs~\cite{chen2024exploring} or retrains graph-specific tokenizers~\cite{chen2024llaga}. The key lies in how to describe TAGs to effectively elicit LLMs’ general knowledge and reasoning abilities for graph-related tasks. Despite their success, \textbf{TAGs introduce new vulnerabilities through the textual modality, exposing novel adversarial attack surfaces.} Given the diverse backbones used for TAGs, there is an urgent need for universal, well-crafted adversarial attacks to expose vulnerabilities and mitigate potential misuse.

% Before TAG models can be deployed in safety-critical domains, they must be rigorously tested with well-designed attacks to uncover potential vulnerabilities and prevent future misuse. 
% Given the diverse architectures of TAG models, a key question arises: \textit{How to design universal adversarial attacks that apply across these architectures to evaluate TAG model security?}

Existing graph attackers typically perturb edges or node features under constrained budgets \cite{wang2024unsupervised,shang2023transferable}, 
but struggle to adapt to TAG models due to their inability to interpret and manipulate textual information.
% However, these approaches struggle to adapt to TAG models due to their inability to interpret and manipulate textual information, resulting in incomplete or suboptimal attacks. 
This aligns with recent findings that graph attackers are less effective in LLM-driven graph learning~\cite{guo2024learning,zhang2024can}.
Furthermore, embedding-based attacks are often impractical and hard to interpret. While recent efforts explore node injection attacks at the textual level \cite{lei2024intruding}, their effectiveness still falls short compared to traditional embedding-space attacks.
To address this gap,  we ask: \textit{How can we design universal adversarial attacks that generalize across architectures to effectively evaluate the security of TAG models?}

Despite its potential, achieving this goal remains challenging.
\textbf{First,} uncovering universal vulnerability patterns across both topological and textual modalities is difficult. The complex interactions between these modalities, along with the diverse ways various backbone architectures encode TAG patterns, make it challenging to identify common weaknesses.
\textbf{Second,} the unknown architecture and parameters of target models often force attackers to operate in the more restrictive black-box setting. This is especially true for commercial LLMs deployed via APIs without internal access, where gradient-based optimization is not feasible. As a result, these constraints hinder the identification of intrinsic vulnerabilities.

Recent advances show that LLMs can act as effective adversarial agents in various domains, including fake news generation~\cite{sun2024exploring} and recommender system~\cite{ning2024cheatagent}, thanks to their ability to generate contextually relevant yet misleading content. These successes motivate us to explore LLMs as powerful attackers in graph learning. 
In this paper, we propose \textsc{BadGraph}, a novel universal adversarial attack on TAG models that harnesses the deep graph reasoning capabilities of LLMs to disrupt the core semantics of TAG.
Specifically, we design a target influencer retrieval module to identify semantically dissimilar candidates, enabling the creation of a shortcut between the target node and the intended misleading label—even in black-box settings without access to internal gradients.
Subsequently, to enable a universal TAG attack, we elicit the LLM’s deep understanding of graph patterns to jointly guide topological and textual perturbations. Notably, our retrieval module facilitates LLM-based edge perturbation in a highly resource-efficient manner.
The attacked TAG can be applied to any target graph learning backbone, including both GNN-based and LLM-based reasoners.
Our key contributions can be summarized as follows:

% \begin{itemize}[leftmargin=*]
\begin{itemize}[leftmargin=*]
\item 
To the best of our knowledge, this is the first universal TAG adversarial attack that generalizes across diverse graph learning models. Our findings reveal that the growing use of TAGs brings new security risks, highlighting the need for a comprehensive evaluation of model robustness.
\item 
% We propose \textsc{BadGraph}, a novel LLM-powered TAG attacker that efficiently leverages the reasoning capabilities of LLMs across both topological and textual modalities.
We propose \textsc{BadGraph}, a novel LLM-powered TAG attacker that efficiently leverages the adversarial capabilities of LLMs across both topological and textual modalities.
\item 
Extensive experiments across diverse TAG backbones demonstrate that \textsc{BadGraph} is both effective and generalizable.  
Moreover, the proposed Cross-Modal Shortcut Theory offers theoretical insight into its stealthiness and efficacy.
\end{itemize}

\section{Related Work}

\subsection{Graph Adversarial Attacks}
Identifying potential vulnerabilities and enhancing the security of GNNs are essential for developing trustworthy graph learning systems~\cite{Zhang_Wu_Yuan_Pan_Tong_Pei_2022}.  
Existing GNN attacks can be broadly classified into two categories: \emph{test-time evasion attacks} and \emph{training-time poisoning attacks}.  
Evasion attacks manipulate the graph structure or node features at inference time to mislead model predictions~\cite{zugner2018adversarial,ma2020towards,wang2022bandits,wang2023turning,wang2024efficient}.  
For example, Nettack proposes a greedy-based edge perturbation strategy \cite{zugner2018adversarial}, while   \cite{li2024graph} extends black-box evasion attacks to the explainability setting.  
In contrast, poisoning attacks perturb the graph during training to embed backdoors or degrade model robustness~\cite{wang2019attacking,zhang2021backdoor,wang2023turning,dai2024semantic,liu2025revisiting,alom2025gottack}.  
SBAC injects semantic trigger nodes to generate poisoned samples \cite{dai2024semantic}, whereas EPD quantifies perturbation importance and distinguishes benign augmentation from adversarial manipulation \cite{liu2025revisiting}.  
GOttack reveals the weaknesses of existing defenses by efficiently compromising GNNs through orbit-based structural manipulations~\cite{alom2025gottack}.  
However, existing works overlook the emerging text-based attack interfaces in the LLM era, which are crucial for preventing  misuse of TAGs in realistic settings.

% This calls for universal adversarial attacks that can adapt to diverse downstream backbones and ensure the safe use of TAG models.

% \cite{lei2024intruding} employed LLMs to elevate embedding-space attacks to the text level.

\subsection{LLM-Generated Attacks}
The utilization of LLMs~\cite{ao2025lightprof,wang2024can}, with their human-like understanding and ability to generate contextually relevant yet misleading content, introduces new capabilities in adversarial attack generation~\cite{sun2024exploring,ning2024cheatagent}. Understanding these new attacks is key to finding AI vulnerabilities and improving their security~\cite{qisafety}.
For example, PromptAttack shows that carefully crafted adversarial prompts can induce a target LLM to produce harmful outputs that deceive itself~\cite{xu2023llm}.
Similarly, PAIR leverages an attacker LLM to iteratively generate jailbreak prompts for a separate target LLM without human intervention~\cite{chao2023jailbreaking}.
 Attack-in-the-Chain uses chain-of-thought prompting to craft attacks against neural ranking models~\cite{liu2025attack}.
CheatAgent leverages an LLM agent to generate targeted adversarial perturbations for attacking recommender systems~\cite{ning2024cheatagent}.
Recently, WTGIA explores text-level injection attacks in the graph domain and finds that their effectiveness decreases as text interpretability increases, making them less effective than embedding-based attacks ~\cite{lei2024intruding}. This highlights the need for further research on the potential  LLM-based attacks in graph learning.
\section{Background and Preliminaries}
\paratitle{Text-attributed Graphs.}
A TAG is defined as $\mathcal{G} = (V, E, S)$, where $V$ is the set of nodes, $E$ is the set of edges, and $S$ denotes node-level textual information. The adjacency matrix of the graph $\mathcal{G}$ is denoted as $A \in \mathbb{R}^{|V| \times |V|}$, where $A_{ij} = 1$ if nodes $v_i$ and $v_j$ are connected, otherwise $A_{ij} = 0$.
In this work, we focus on the node classification task on TAGs. Specifically, each node $v_i$ corresponds to a label $y_i$ that indicates which category the node $v_i$ belongs to. When attacking GNNs serving as reasoners, we encode the text $S$ into the node feature matrix $X = \{x_1, \dots, x_{|V|}\}$ using various encoding techniques to train the target GNNs, where $x_i \in \mathbb{R}^d$. 
% Given some labeled nodes $V_L \subset V$, the goal is training a GNN $f(A, X)$ to predict the labels of the remaining unlabeled nodes $V_U = V \setminus V_L$.

\paratitle{Attacker’s Knowledge and Capability.} 
We study a  black-box evasion setting, i.e., the attacker has no access to model architectures, parameters, training procedures, or text encoders, and is limited to test-time, architecture-agnostic universal attacks on TAGs.
% To realize the attack, we propose an LLM-driven TAG attacker that leverages the LLM's strong reasoning capabilities and its broad knowledge of graph pattern.

\paratitle{Attacker's Goal.}
The attacker's objective is to generate a perturbed graph $\mathcal{G}'=(A',S')$ by modifying the adjacency $A$ and node text $S$ under a limited budget (e.g., $\|\mathcal{G}' - \mathcal{G}\| \le \Delta$), so that the target model outputs a predefined incorrect label $Y'$. 
We focus on the challenging setting of universal, black-box attacks, where the adversarial graph $\mathcal{G}'$ is required to transfer across heterogeneous backbones with substantial differences in architecture and inference, including 
(1) a GNN reasoner $f^{\mathrm{GNN}}_{\theta^{(k)}}(A,S)$ and 
(2) an LLM reasoner $f^{\mathrm{LLM}}_{\theta^{(k)}}\!\big(\mathcal{T}_{t}(A,S)\big)$, 
with $\theta^{(k)}$ denoting the $k$-th model instance and $\mathcal{T}_{t}$ a task-specific prompt. 
Formally, given $\mathcal{G}'=(A',S')$, the attacker enforces:
\begin{equation}
\begin{aligned}
& f^{m}_{\theta^{(k)}}(A',S') \;\rightarrow\; Y', 
\quad m \in \{\mathrm{GNN}, \mathrm{LLM}\}, \\
& \text{s.t. } \|A'-A\|_0 \le b_A,\;\; \|S'-S\|_0 \le b_S,
\end{aligned}
\end{equation}
where $b_A$ and $b_S$ bound the maximum number of structural and textual edits, respectively.

% \quad\text{or}\quad
% f^{\mathrm{LLM}}_{\phi}\big(\mathcal{T}_{t}(A',S')\big) \rightarrow Y'.
% \end{equation}

% $f^{\text{GNN}}_\theta(A', S') \rightarrow Y'$ or $f^{\text{LLM}}_\phi(\mathcal{T}_{t}(A', S')) \rightarrow Y'$.

% $t$ denotes a textual prompt that explicitly defines the downstream task as node classification.
% \subsection{Candidate Trigers}

 \begin{figure*}
  \includegraphics[width=0.75\textwidth]{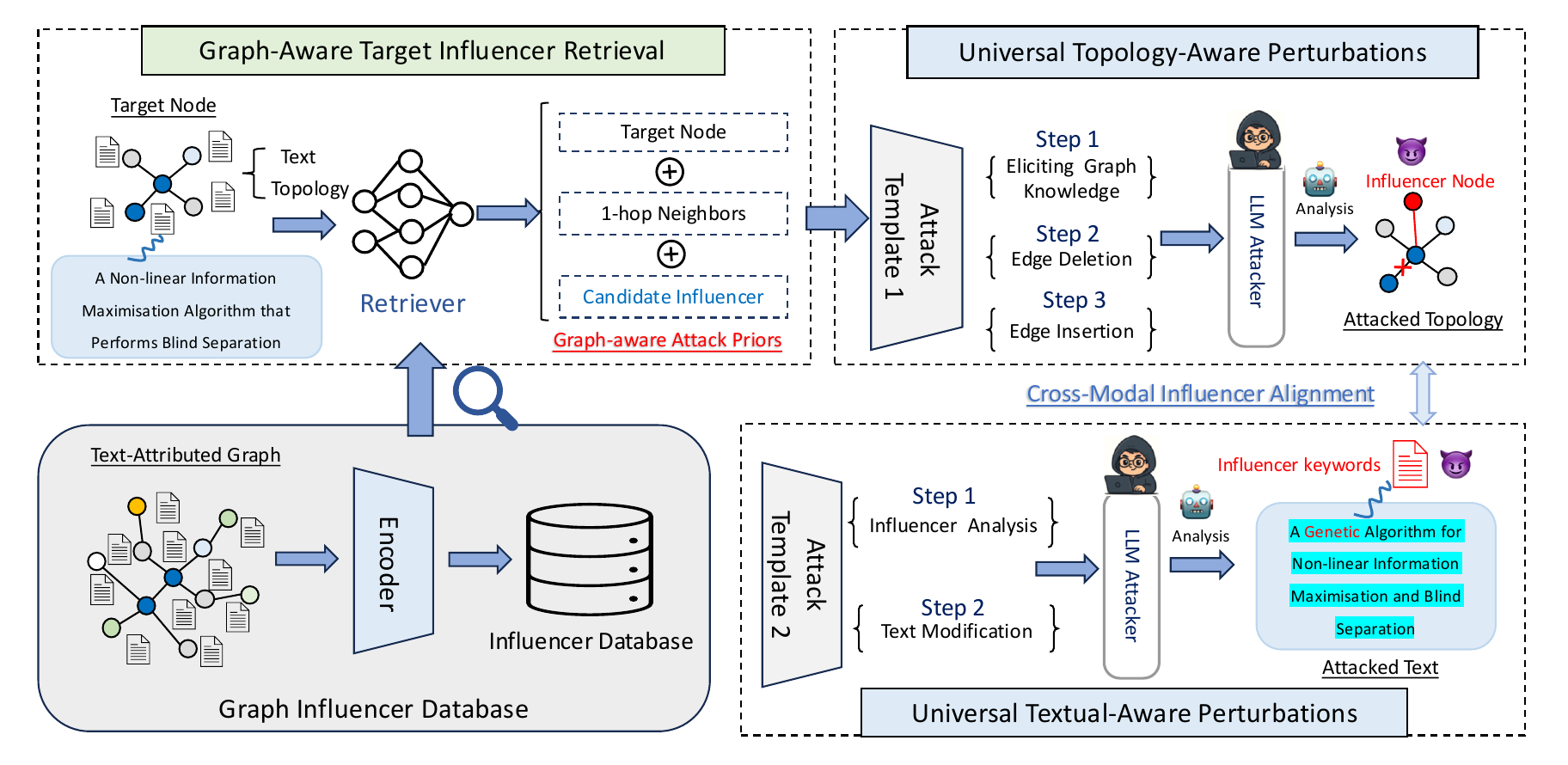}
  \caption{An overview of the proposed \textsc{BadGraph} framework.}
  \label{model}
\end{figure*}
\section{Proposed Framework}
% \textsc{BadGraph} aims to generate attacked graph through LLMs with Chain-of-thought (COT), as illustrated in Figure \ref{model}. Our method involves three main parts: 1) GNN-based Retrieval, 2) Step-by-Step Structural attack, 3) Step-by-Step Textual attack.

% In this paper, we aim to construct a universal, model-agnostic perturbation for TAGs that can be applied to any target model, regardless of its architecture. This perturbation is crafted to consistently degrade the performance of a wide range of graph learning models. To this end, we propose \textsc{BadGraph}, a method that progressively guides the LLM’s understanding of graph patterns to launch universal topological and textual adversarial attacks. The overall pipeline is illustrated in Figure~\ref{model}.

In this paper, we propose \textbf{\textsc{BadGraph}}, a novel LLM-driven adversarial attack tailored for TAGs. \textsc{BadGraph} progressively guides an LLM to discover transferable perturbations over topology and text. The retrieval-then-reasoning pipeline is illustrated in Figure~\ref{model}.

% \textsc{BadGraph} progressively guides an LLM to generate transferable perturbations over topology and text, yielding model-agnostic attacks across heterogeneous TAG backbones, as illustrated in Figure~\ref{model}.

\subsection{Graph-Aware Target Influencer Retrieval}

% Given a test-time graph $\mathcal{G}$, our goal is to leverage an LLM's graph understanding to construct a perturbed graph $\mathcal{G}'=(A',S')$ by editing the adjacency $A$ and node text $S$ so that a target model $f^m_{\theta^{(k)}}(A',S')$ outputs a predefined, incorrect label $Y'$. 
% The main challenges lie in efficiency and the absence of graph-specific attack priors under black-box constraints. Without gradient access, topology-based attacks (e.g., edge additions) would require querying the LLM across all node pairs, resulting in $\mathcal{O}(N^2)$ complexity and prohibitive cost. Moreover, the lack of domain priors makes it difficult for LLMs to localize effective perturbation regions within the graph.
To leverage LLM-based graph reasoning for constructing a perturbed graph $\mathcal{G}'=(A',S')$, such that the target graph learning model $f^m_{\theta^{(k)}}(A',S')$ is misled to produce a predefined incorrect label $Y'$.
The main challenges are twofold:
(1) efficiency under black-box constraints—without gradient access, topology-based attacks (e.g., fake edge additions) necessitate querying all node pairs via the LLM, leading to an $\mathcal{O}(V^2)$ query cost; and
(2) the absence of graph priors, which hinders the LLM from deriving reliable attack cues and accurately localizing effective perturbation regions.
% The main challenges lie in (1) efficiency under black-box constraints and (2) the absence of graph-specific attack priors. 
% Without gradients, topology-based attacks (e.g., edge additions) require querying the LLM for all node pairs—incurring $\mathcal{O}(N^2)$ complexity and prohibitive cost. 
% Meanwhile, the lack of graph priors hinders the LLM’s ability to localize effective perturbation regions within the graph.

To overcome these challenges, we introduce a Target Influencer Retrieval module that narrows the search space to a compact set of \emph{influencer nodes}, thereby avoiding exhaustive pairwise prompting. 
The retrieved nodes act as graph-aware priors, supplying explicit shortcut paths to the incorrect label.
Formally, we denote the influencer of a target node $v_i$ as $v_i^{\mathrm{influencer}}$, whose label $y_{v_i^{\mathrm{influencer}}}$ can misdirect the target model’s prediction on $v_i$.
% Formally, let an influential node of the target node $v_i$ be the one whose label can misdirect the model’s prediction on $v_i$.
We then encode the input graph $\mathcal{G}$ into node embeddings $Z \in \mathbb{R}^{|\mathcal{V}|\times d}$ :

\begin{equation}
    Z = \operatorname{GraphEncoder}(A, X),
\end{equation}
where $Z$ serves as a model-agnostic influencer database for retrieval.
The graph encoder can be instantiated with any lightweight or expressive GNN (e.g., GCN, GAT); empirically, more expressive encoders lead to stronger and more transferable perturbations.

% Despite the strong reasoning capabilities of LLMs, adapting them for TAG attacks is non-trivial due to two key challenges. 
% First, the cross-modal nature of TAGs requires perturbations that consistently anchor to the same misleading label, inducing semantic and structural shortcuts with minimal intervention. 
% Second, the high inference latency of LLMs imposes strict efficiency constraints: topology-based attacks, such as edge additions, would demand prompting over all node pairs, which is computationally prohibitive. 
% To address these challenges, we propose a retrieval-first design that selects a compact set of structurally salient \emph{influencer nodes}. This targeted retrieval drastically reduces the search space, enabling focused and low-cost perturbations.

% Specifically, we define the influencer for each target node $v_i$ as $v^{\text{inf}}_{i}$, which is a node that can mislead any target model into predicting the label of $v_i$ as that of its influencer, i.e., $y^{v^{\text{inf}}_{i}}$. To achieve this goal, we propose a target influencer retrieval module, which encodes nodes in the graph $\mathcal{\mathcal{G}}$ into embeddings $Z$ using a simple encoder.

% \begin{equation}
%     {Z} = \operatorname{GraphEncoder}\left(A, X\right),
% \end{equation}
% where $Z$ serves as our influencer database.

For each target node $v_i$ with embedding $z_i$, we retrieve a compact candidate influencer set $\mathcal{I}_{v_i}$ that are semantically distant from $v_i$. Let $d_{\cos}(z_i,z_v)$
denote cosine-based dissimilarity; the retriever selects the top-$K$ nodes with the largest dissimilarity to $v_i$:
\begin{equation}
\mathcal{I}_{v_i} \;=\; \operatorname{Retriever}\big(\{d_{\cos}(z_i,z_v)\}_{v\in\mathcal{V}\setminus\{v_i\}},\,K\big),
\end{equation}
where $\mathcal{I}_{v_i}$ forms the candidate pool for subsequent LLM-driven adversarial perturbations. 
By concentrating LLM reasoning on a small, high-quality subset,  our retrieval-first design drastically reduces query cost and injects graph-aware attack priors.

\subsection{LLM-Driven Cross-Modal TAG Perturbation}

LLMs have recently exhibited a surprising degree of graph reasoning intelligence.
Building on this insight, we investigate how such emergent reasoning can be harnessed to mount adversarial attacks on TAGs. 
A central challenge remains: the dual nature of TAGs requires that topological and textual edits be tightly aligned; directional misalignment substantially undermines attack effectiveness.
% To ensure such cross-modal coherence,  for each target node $v_i$ (with retrieved influencers $\mathcal{I}_{v_i}$), we jointly generate anchor-based perturbations over graph topology and node text, forming consistent shortcuts that steer the model toward the intended misleading label.
To ensure cross-modal coherence, for each target node $v_i$ with retrieved influencers $\mathcal{I}_{v_i}$, we jointly generate anchor-based topological and textual perturbations, forming aligned shortcuts that steer the model toward the target misleading label.

% LLMs have shown increasing capability in understanding and reasoning over complex graphs.
% In this paper, we explore how LLMs can leverage their understanding of graph patterns to guide topological attacks—by removing critical edges around a target node and adding misleading ones. Achieving this is challenging, primarily because:
% (1) it is non-trivial to elicit the LLM's understanding of graph patterns with only limited prior knowledge; and (2) performing edge prediction with LLMs is inherently difficult due to the $\mathcal{O}(N^2)$ complexity of pairwise edge inference.

% Prior work has demonstrated that LLMs possess a certain capability to understand graph structural information. However, in the context of structural perturbation, directly using LLMs to attack a target node typically requires traversing the entire graph, resulting in a time complexity of O(N) per attack. 
 % We query the LLM in an "open-ended" manner, i.e., prompt LLM to make perturbations.
% To incorporate graph structural information into the prompt and generate harmful prompts that enable the LLM to reason and produce adversarial examples, we design structure-aware attack prompts consisting of two main components: Instruction and Attack Step Guidance.

\subsubsection{Universal Topology-Aware Perturbations} 
Topology manipulation is central to constructing adversarial shortcuts on TAGs. 
Here, the LLM is prompted to produce budget-constrained structural perturbations—edge removals and additions—guided by the retrieved influencers. 
These topology-aware edits jointly reshape the local connectivity of the target node.

\paratitle{Eliciting  Graph Knowledge.} We first prompt the LLM to extract graph-relevant cues from limited prior information, thereby eliciting the LLM’s implicit knowledge of graph patterns.
Specifically, for each target node $v_i$, we construct a structured prompt that directs the LLM to reason over its local adjacency $\mathcal{N}_{v_i}$ and node text $S_i^{(1)}$:

\begin{equation}
\mathcal{R}_{i} 
= \mathcal{T}_{\text{graph}}\!\big(v_i,\, \mathcal{N}_{v_i},\, S_i^{(1)}\big),
\end{equation}
where $\mathcal{T}_{\text{graph}}(\cdot)$ instructs the LLM to infer relational cues between $v_i$ and its neighbors, yielding a reasoning summary $\mathcal{R}_{i}$ that serves as the foundation for subsequent perturbation generation.

\begin{tcolorbox}[
  colback=gray!5!white,
  colframe=gray!40!black,
  boxrule=0.2pt,
  left=2pt,right=2pt,top=2pt,bottom=2pt,
  arc=2pt,
]
% \textbf{\# Eliciting  Graph Knowledge}\\
% A graph describes the complex relationships between nodes. For a target node, its neighboring nodes form a set that is highly related to the target node from a certain perspective. In a citation network, given a target node \{Target Node\}, the set of its highly related neighboring nodes is: \{1-hop Neighbors\}.
% Please follow the steps below to assist in performing an accurate correlation analysis between nodes. \\
\textbf{Step 1}: Analyze the target node and its neighboring set.\\
\textcolor{blue}{Summarize why the nodes in the neighboring set are adjacent to the target node.} Be sure to highlight the most prominent factors that guide their strong correlation. 
\end{tcolorbox}
% \vspace{-\baselineskip}
% \vspace{-0.2\baselineskip}

\paratitle{Edge Deletion Guidance.}
Building upon the structural understanding elicited in the Step 1, we guide the LLM to identify and attack a critical edge of a target node $v_i$ via stepwise reasoning.
% The objective is to degrade the target node  $v_i$'s local structural support by removing its most supportive neighbor.
Formally, given the 1-hop neighborhood $\mathcal{N}_{v_i}$, the LLM performs a stepwise reasoning process to select the most semantically relevant neighbor as the deletion target:
\begin{equation}
v_i^{\mathrm{del}} = \arg\max_{v \in \mathcal{N}_{v_i}} \mathrm{LLM}_{\mathrm{rel}}(v_i, v).
\end{equation}
The removal of edge $(v_i, v_i^{\mathrm{del}})$ reduces the local structural reinforcement for $v_i$ and weakens the evidence supporting its original label. 

% Formally, the LLM examines the 1-hop neighborhood $\mathcal{N}_{v_i}^{(1)}$ and nominates the deletion target:
% \begin{equation}
% v_i^{\mathrm{del}} \;=\; \operatorname{Select}_{\mathrm{LLM}}\!\big(v_i,\, \mathcal{N}_{v_i}^{(1)}\big),
% \end{equation}
% where $\operatorname{Select}_{\mathrm{LLM}}$ is a LLM-driven selection operator (optionally realized as a ranking). 
% The removal of edge $(v_i, v_i^{\mathrm{del}})$ reduces the local structural reinforcement for $v_i$ and weakens the evidence supporting its original label. 

% \noindent
% \colorbox{gray!10}{%
%   \parbox{\dimexpr\linewidth-2\fboxsep}{%
%     \vspace{4pt}
%      \textbf{\# Edge Deletion Guidance}\\
% \textbf{Step 2}: From the neighboring set, choose the node that is \textcolor{blue}{most relevant to the target node}. Let's break it down step by step to ensure we accurately evaluate the correlation. 
%     \vspace{4pt}
%   }%
% }
% \vspace{-\baselineskip}
\begin{tcolorbox}[
  colback=gray!5!white,
  colframe=gray!40!black,
  boxrule=0.2pt,
  left=2pt,right=2pt,top=2pt,bottom=2pt,
  arc=2pt,
]
% \textbf{\# Edge Deletion Guidance}\\
\textbf{Step 2}: From the neighboring set, choose the node that is \textcolor{blue}{most relevant to the target node}. Let's break it down step by step to ensure we accurately evaluate the correlation. 

\end{tcolorbox}
\paratitle{Edge Insertion Guidance.}
To further distort the neighborhood structure, we leverage the retrieved influencer candidates $\mathcal{I}_{v_i}$ as potential targets for edge addition; this retrieval set provides principled attack evidence—its semantic gap to $v_i$.
The LLM analyzes the semantic relationships within $\mathcal{I}_{v_i}$ and selects the least semantically related node as the addition target:
\begin{equation}
v_i^{\mathrm{add}} = \arg\min_{v \in \mathcal{I}_{v_i}} \mathrm{LLM}_{\mathrm{rel}}(v_i, v).
\label{add}
\end{equation}
Adding the edge $(v_i, v_i^{\mathrm{add}})$ introduces a misleading connection to a semantically distant node, constructing an adversarial shortcut that biases the model toward the influencer’s label. 
This process reduces the edge prediction complexity from $\mathcal{O}(N^2)$ to $\mathcal{O}(N)$, as the LLM focuses reasoning on a compact influencer set.

% To address the second challenge, we leverage the retrieved candidate influencer set $\mathcal{I}_{v_i}$ as potential targets for edge addition. Aiming to disrupt the target node’s local structure by adding a misleading, unrelated neighbor, we prompt the LLM to analyze the set $\mathcal{I}_{v_i}$ and select the least semantically related node as the edge addition target, denoted as $v^{\text{add}}_{i}$. This reduces the complex edge prediction task $\mathcal{O}(N^2)$ to a simpler selection process $\mathcal{O}(N)$, as illustrated below.

% Attack Step Guidance: This part guides the model through a step-by-step reasoning process. It instructs the LLM to analyze the relationship between the $v_T$ and its 1-hop neighbors, identify the most semantically relevant neighboring node denoted as $v_{delete}$, and the most irrelevant node from a candidate set denoted as $v_{add}$. The goal is to disrupt the local structural information of the target node by removing the most supportive neighbor and adding a misleading, unrelated neighbor.

% \noindent
% \colorbox{gray!10}{%
%   \parbox{\dimexpr\linewidth-2\fboxsep}{%
%     \vspace{4pt}
%     \textbf{\# Edge Insertion Guidance}
    
%     \textbf{Step 3}: Based on the inferred prominent factors from Step 1, exclude the node from the following Candidate List that is \textcolor{blue}{least related to the target node}e and analyze why.
    
%     \textbf{Candidate List}: \{Candidate Influencers\}
%     \vspace{4pt}
%   }%
% }
% \vspace{-\baselineskip}
\begin{tcolorbox}[
  colback=gray!5!white,
  colframe=gray!40!black,
  boxrule=0.2pt,
  left=2pt,right=2pt,top=2pt,bottom=2pt,
  arc=2pt,      % 减少盒子内部下边距
]
% \textbf{\# Edge Insertion Guidance}

\textbf{Step 3}: Based on the inferred prominent factors from Step 1, exclude the node from the following Candidate List that is \textcolor{blue}{least related to the target node} and analyze why. 

% Candidate List: \{Candidate Influencers \}. 
\end{tcolorbox}

\paratitle{Unified Formulation.}
The resulting topology perturbation for node $v_i$ can thus be represented as:
\begin{equation}
\Delta A_i = \big\{ (v_i, v_i^{\mathrm{add}}) \big\} \cup \big\{ (v_i, v_i^{\mathrm{del}}) \big\},
\qquad 
\|\Delta A_i\|_0 \le b_{A_i},
\end{equation}
where $b_{A_i}$ is the topology perturbation budget.  
This unified formulation allows the LLM to perform topology attacks within a consistent reasoning framework, yielding universal and transferable topology-aware perturbations.  
Empirically, we set $b_{A_i}=2$, corresponding to one edge addition and one edge deletion per target node.

\subsubsection{Universal Text-Aware Perturbations}
LLMs have proven effective at crafting adversarial text in NLP.
Inspired by this, we harness LLMs' human-like understanding of semantics and graph context to generate targeted textual perturbations for TAGs, providing a principled, text-centric adversarial interface. 
The main challenge is to induce model misclassification with minimal, stealthy edits: textual changes must be small yet directionally consistent with topology edits so as to form coherent cross-modal shortcuts.

% LLMs have demonstrated increasing effectiveness in performing textual attacks within natural language processing tasks. Inspired by this, we investigate how LLMs can leverage their human-like understanding of both textual semantics and graph topology to guide effective textual attacks on TAGs, thereby providing a principled, text-based interface for generating adversarial perturbations. The key challenge lies in inducing misclassification by the LLM with minimal textual perturbation, so as to keep the attack stealthy and less detectable.

% We observe that directly instructing the LLM to modify textual information often leads the model to retain key features that are representative of the original category, which fails to meet the adversarial objective. 
% To address this, we introduce $v_{add}$ into the prompt which are analyze by LLM in structural attack. The textual attack prompt is divided into two parts: Attack Objective and Attack Guidance.
% We aim to generate an adversarial sample that maintains a high degree of textual similarity to the original, while misleading the LLM into an incorrect classification. The adversarially generated text by the LLM should be classified as $y^{v^{\text{add}}_{i}}$.

\paratitle{Influencer Analysis Guidance.} 
% In this case, we still use the target influencer identified during the structural attack as the source of perturbation. To encourage the misclassification of the target node into the class of the target influencer, we introduce $v^{\text{add}}_{i}$ into the prompt and instruct the LLM to identify the most important keywords that reflect its category $y^{v^{\text{add}}_{i}}$. 
To maintain cross-modal consistency, we follow the same influencer $v_i^{\mathrm{add}}$ selected in the structural attack (Eq.~\eqref{add}) as the textual perturbation source, steering each target node $v_i$ toward the class of its influencer $y_{v_i^{\mathrm{add}}}$. 
Formally, the LLM is prompted to extract a representative keyword from the influencer node $v_i^{\mathrm{add}}$:
\begin{equation}
k_{v_i} = \mathrm{LLM}_{\text{key}}\!\big(v_i^{\mathrm{add}},\, S_{v_i^{\mathrm{add}}}\big),
\end{equation}
where $k_{v_i}$ denotes the category-indicative keyword extracted from the influencer’s text $S_{v_i^{\mathrm{add}}}$, 
which serves as a semantic anchor for subsequent textual perturbations on $v_i$.

% \noindent
% \colorbox{gray!10}{%
%   \parbox{\dimexpr\linewidth-2\fboxsep}{%
%     \vspace{6pt}
%     \textbf{\#Influencer Analysis Guidance}

% A graph describes the complex relationships between nodes. For node classification, in citation network, there are \{category size\} category: \{category\}

% Please follow the steps below to assist in generating a new sentence. 

% \textbf{Step 1}: Given the target node titled \{$v^{\text{add}}_{i}$\}, \textcolor{blue}{identify one keyword that reflects its category}.
%     \vspace{6pt}
%   }%
% }
% \vspace{-\baselineskip}
\begin{tcolorbox}[
  colback=gray!5!white,
  colframe=gray!40!black,
  boxrule=0.2pt,
  left=2pt,right=2pt,top=2pt,bottom=2pt,
  arc=2pt,
]
% \textbf{\#Influencer Analysis Guidance}

% A graph describes the complex relationships between nodes. For node classification, in citation network, there are \{category size\} category: \{category\}

% Please follow the steps below to assist in generating a new sentence. 

\textbf{Step 1}: Given the target node titled \{$v^{\text{add}}_{i}$\}, \textcolor{blue}{identify one keyword that reflects its category}.

\end{tcolorbox}

% the content of the added node identified from the structure attack, and require the LLM to infer the category $y_{v_{add}}$. The generated adversarial sample should be classified as $y_{v_{add}}$., the content of the added node identified from the structure attack, and require the LLM to infer the category $y_{v_{add}}$. The generated adversarial sample should be classified as $y_{v_{add}}$.
% \paratitle{Text Modification Guidance.}
% Then, based on the keywords of the influencer obtained in the previous step, we guide the LLM to perturb the target node's text so that it is classified into the same category as $y^{v^{\text{add}}_{i}}$.
% Two core perturbation constraints are introduced: the generated adversarial sample must retain some words from the original input; it must incorporate the keyword from $v^{\text{add}}_{i}$, so that the adversarial sample aligns with $y^{v^{\text{add}}_{i}}$.

\paratitle{Text Modification Guidance.}
Building on the extracted influencer keyword $k_{v_i}$, we prompt the LLM to subtly revise the target node’s text so that it aligns with the influencer’s category $y_{v_i^{\mathrm{add}}}$. 
To ensure both stealth and cross-modal coherence, two constraints are imposed:  
(1) retain partial lexical content to preserve linguistic fluency and adversarial stealth; and  
(2) integrate the influencer keyword $k_{v_i}$ form Eq.~\eqref{key}, ensuring directional semantic drift toward $y_{v_i^{\mathrm{add}}}$.
Formally, the adversarially modified text is generated as:
\begin{equation}
S_i' = \mathrm{LLM}_{\text{text}}\!\big(S_i,\, k_{v_i}\big),
\quad \text{s.t. } \|S_i' - S_i\|_0 \le b_{S_i},
\label{key}
\end{equation}
where ${S_i}$ denotes the textual perturbation budget.  
This formulation allows the LLM to synthesize minimal yet semantically effective text edits, achieving coherent alignment between structural and textual perturbations under strict black-box constraints.

% \noindent
% \colorbox{gray!10}{%
%   \parbox{\dimexpr\linewidth-2\fboxsep}{%
%     \vspace{6pt}
%     \textbf{\#Text Modification Guidance}
    
% \textbf{Step 2}: Given the paper P1 titled  \{$s_{v_i}$\}, your task is to generate a new paper by modifying P1 title so that it meets the following requirements:

% 1. It must \textcolor{blue}{retain some of the original words} from the P1 title.

% 2. It should \textcolor{blue}{include the keyword} identified in Step 1 and be aligned with the target node \textcolor{blue}{category determined in Step 1}.
%     \vspace{6pt}
%   }%
% }
% \vspace{-\baselineskip}
\begin{tcolorbox}[
  colback=gray!5!white,
  colframe=gray!40!black,
  boxrule=0.2pt,
  left=2pt,right=2pt,top=2pt,bottom=2pt,
  arc=2pt,
]
% \textbf{\#Text Modification Guidance}

\textbf{Step 2}: Given the paper P1 titled  \{$S_i$\}, your task is to generate a new paper by modifying P1 title so that it meets the following requirements:  \\
1. It must \textcolor{blue}{retain some of the original words} from the P1 title.\\
2. It should \textcolor{blue}{include the keyword} identified in Step 1 and be aligned with the target node \textcolor{blue}{category determined in Step 1}.
\end{tcolorbox}

% \paratitle{In summary.}
Departing from gradient-driven or retraining-based attacks, \textsc{BadGraph} unifies structural and textual attacks within a LLM reasoning pipeline.  
% By anchoring both modalities to a shared influencer, i
It achieves directionally aligned and transferable attacks at low cost, requiring only two LLM queries per target node.

\section{Theoretical Foundations of \textsc{BadGraph}.}

A key question arises regarding the stealthiness of \textsc{BadGraph}:  
how can it remain effective while inducing only minimal disruption?  
To resolve this paradox, we establish a unified theoretical framework with two complementary components:  
(1) a \emph{homophily-preserving bound}, explaining why \textsc{BadGraph} maintains stealthiness under sparse perturbations; and  
(2) a \emph{cross-modal shortcut theory}, revealing how such subtle perturbations achieve strong attack efficacy.

\subsection{Homophily-Preserving Bound}
The adversary jointly perturbs topology and text to maximize the probability of misclassification:
\begin{equation}
\max_{\|A'-A\|_0 \le b_A,\;\|S'-S\|_0 \le b_S}
\mathbb{P}\!\left(Y' \mid A', S'; f^{m}_{\theta^{(k)}}\right).
\end{equation}
% where $b_A$ and $b_S$ denote the structural and textual budgets, respectively.  
% Under this formulation, \textsc{BadGraph} enforces sparsity in topological edits and semantic alignment in textual modifications through LLM-guided reasoning.

Let $x_i = \Phi(S_i)$ denote the textual embedding, where we assume that the encoder $\Phi$ is $L_\Phi$-Lipschitz continuous. 
A bounded textual perturbation $ \|S' - S\|_0 \le b_{S_i}$ thus induces a semantic deviation 
$\|x_i' - x_i\|_2 \le L_\Phi \tau(b_{S_i})$, 
where $\tau(b_{S_i})$ quantifies the embedding drift.  
Similarly, the relative edge perturbation ratio is $\Delta_E = \|A'-A\|_0 / |E| \ll 1$ .  
Then, following the node-centric definition of homophily~\cite{chen2022understandingimprovinggraphinjection},
the change in global homophily satisfies:
\begin{equation}
|H(G') - H(G)| \le C_1 \Delta_E + C_2 L_\Phi \tau(b_S),
\end{equation}
where $C_1, C_2 > 0$ depend on graph smoothness and the continuity of the similarity metric.  
Because both $\Delta_E$ and $\tau(b_S)$ remain small under \textsc{BadGraph}’s 
budget-constrained and semantically aligned perturbations, 
the resulting global homophily variation is provably bounded---offering a formal guarantee of stealthiness.  In section \ref{sec:empirical_stealthiness}(Observation 2) further corroborate this homophily stability.

\subsection{Cross-Modal Shortcut Theory}
% Formally, given a target node $v_i$ and its retrieved influencer node $v_i^{\mathrm{add}}$ with label $y_{v_i^{\mathrm{add}}} \neq y_{v_i}$,
% the attacker seeks to jointly perturb structure and text to maximize the misclassification probability.
% \begin{equation}
% (A_i^{\prime *}, S_i^{\prime *}) 
% = \arg\max_{\|A'-A\|_0 \le b_A,\;\|S'-S\|_0 \le b_S}
% \mathbb{P}\!\left(Y' \mid A', S';\, f^m_{\theta^{(k)}}\right),
% \end{equation}
% where $b_A$ and $b_S$ denote the budgets for structural and textual edits, respectively.
Although the joint optimization $(A_i^{\prime *}, S_i^{\prime *})$ is intractable under black-box constraints, a capable LLM can \emph{implicitly approximate} this reasoning objective:
\begin{equation}
(A_i^{\prime *}, S_i^{\prime *}) 
\approx 
\operatorname{LLM}\!\big(v_i,\, \mathcal{N}_{v_i},\, S_i^{(1)},\, \mathcal{I}_{v_i}\big),
\end{equation}
yielding two coordinated perturbations:
\begin{itemize}
    \item a \emph{structural shortcut} $\delta_A$ that connects the target node $v_i$ to its influencer $v_i^{\mathrm{add}}$, and  
    \item a \emph{textual shortcut} $\delta_S$ that semantically aligns $S_i$ toward $S_{v_i^{\mathrm{add}}}$,
\end{itemize}
both coherently anchored on the same influencer node $v_i^{\mathrm{add}}$.
This alignment induces a directional shift in the latent space:
\[
z(v_i') \approx z(v_i^{\mathrm{add}}) 
\quad\Rightarrow\quad
f\!\big(z(v_i')\big) \rightarrow y_{v_i^{\mathrm{add}}},
\]
where $z(\cdot)$ is the TAG encoder.
By anchoring both modalities to the same influencer and label $y_{v_i^{\mathrm{add}}}$,  
LLM-guided reasoning enforces \emph{cross-modal coherence}, 
allowing the structural and textual shortcuts to reinforce each other and form a unified adversarial pathway.

Consequently, the joint perturbation exhibits a \emph{synergistic effect}:
\begin{equation}
\Delta_{\mathrm{joint}} 
\approx \Delta_{\delta_A,\delta_S} 
> \Delta_{\delta_A} + \Delta_{\delta_S},
\end{equation}
This indicates that aligned cross-modal perturbations substantially outperform isolated attacks.  
As shown in Table~\ref{tab:different_prompt} (Anchor Mis), disrupting this alignment sharply lowers success rates, underscoring the importance of cross-modal coherence for effective transfer.

% \paratitle{Theoretical Implication.}
% This cross-modal coordination explains why even small, budget-constrained perturbations yield large adversarial impact.
% By aligning topology- and text-level perturbations along a shared semantic–structural direction,
% \textsc{BadGraph} constructs a coherent shortcut that simultaneously amplifies attack effectiveness and enhances transferability—while keeping global homophily nearly invariant.
% Empirically, our ablation in Table~\ref{tab:ablation_alignment} confirms that misaligning these cross-modal directions significantly reduces attack success, validating the theory.

\section{Experiments}

\subsection{Experimental Settings}

% \subsubsection{Datasets}
\subsubsection{Datasets.}
We conduct extensive experiments on three TAG datasets: Cora \cite{mccallum2000automating}, OGBN-Products, and OGBN-Arxiv \cite{hu2020open}. 
More details and statistics of the dataset can be found in Appendix \ref{appendix:dataset}.

% We conduct extensive experiments on three TAG datasets: Cora \citet{mccallum2000automating}, ogbn-Arxiv, ogbn-Products \cite{hu2020open}. For the Cora and ogbn-Arxiv datasets, we formatted the raw texts by extracting the title. For the ogbn-Products dataset, we formatted the raw texts using the product name; if the name is missing, we used the product description instead. The dataset statistics are provided in Table \ref{datasets}. We follow the data splitting scheme from \cite{zheng2021graph}, using the full splits for each dataset to train the target model. For the attack settings, we use the medium split as target nodes in Cora. In ogbn-Arxiv, we randomly select 1000 nodes from the medium split as target nodes, and in ogbn-Products, we randomly select 1000 nodes from the hard split. Each node is allowed up to 2 edge perturbations. For word embedding techniques, we include TF-IDF and Sentence-BERT (SBERT). We report the average performance over 5 seeds for each result.
% % and standard deviation 
% % Further details, including standard deviations, can be found in Appendix D
% \begin{table}
% \centering
% \caption{Statistics of datasets.}
% \label{datasets}
% \begin{tabular}{lccc}
% \toprule
% Dataset & \#Nodes & \#Edges & \#Classes \\
% \midrule
% Cora & 2,708 & 5,429 & 7 \\
% ogbn-Arxiv & 169,343 & 1,166,243 & 40 \\
% ogbn-Products (subset) & 54,025 & 74,420 & 47 \\
% \bottomrule
% \end{tabular}
% \end{table}
\subsubsection{Target Backbones.} 
% We conduct attacks across a variety of widely used target models built on two different backbone architectures.
% (1) GNN as a Reasoner. In this setting, node features are first generated using text encoders, including the shallow method TF-IDF and the pretrained language model SBERT.
% These features are then fed into different GNNs, such as GCN \cite{kipf2016semi}, GIN \cite{xu2018powerful}, GraphSAGE \cite{hamilton2017inductive}, TAGCN \cite{du2017topology}, and SGCN \cite{wu2019simplifying}, and a robust variant, R-GCN \cite{zhu2019robust}. (2) LLM as a reasoner. In this setting, LLMs are directly used for node classification tasks. Following \cite{chen2024exploring}, we use Mistral-7B \cite{jiang2024mixtral} and DeepSeek \cite{lu2024deepseek} in a zero-shot setting as target backbones, where the model makes predictions based on node textual descriptions and information from the 2-hop neighborhood. More details about the target models shown in the \todo{Appendix}.

We evaluate \textsc{BadGraph} across two backbone paradigms.  
\textbf{(1) GNN Reasoners.} Node texts are first encoded, then processed by GNNs~\cite{kipf2016semi,xu2018powerful,hamilton2017inductive,du2017topology}, and the robust R-GCN~\cite{zhu2019robust}.  
\textbf{(2) LLM Reasoners.} Following~\cite{chen2024exploring}, LLMs  are used for zero-shot node classification~\cite{jiang2024mixtral,lu2024deepseek}, with details in Appendix~\ref{appendix:target_backbone}.

% \textbf{(1) GNN Reasoners.} Node texts are first encoded, then processed by representative GNNs, such as GCN~\cite{kipf2016semi}, GIN~\cite{xu2018powerful}, GraphSAGE~\cite{hamilton2017inductive}, TAGCN~\cite{du2017topology}, and the robust R-GCN~\cite{zhu2019robust}.  
% \textbf{(2) LLM Reasoners.} Following~\cite{chen2024exploring}, Mistral-7B~\cite{jiang2024mixtral} and DeepSeek~\cite{lu2024deepseek} are employed in a zero-shot manner to perform node classification based on textual descriptions and their 2-hop neighborhoods.  
% Implementation details are provided in the  Appendix \ref{appendix:target_backbone}.

\subsubsection{Baselines.}
We compare \textsc{BadGraph} against seven representative graph attack baselines:  
RND, FLIP~\cite{bojchevski2019adversarial}, STACK~\cite{xu2012query}, PGD~\cite{madry2018towards}, NETTACK~\cite{zugner2018adversarial}, SGAttack~\cite{li2023adversarial}, and WTGIA~\cite{lei2024intruding}.  
For comparison, we assume that the attacker has no knowledge of embedding technology.
Additional details are provided in Appendix \ref{appendix:baseline}.
% For baselines, we use GCN as the surrogate model.
% For WTGIA, we select the best-performing attack variant reported in the original paper. Specifically, we utilize the perturbed graph generated by TDGIA \cite{zou2021tdgia} after training, and employ Llama3-8B \cite{touvron2023llama} to convert the embedding information of the fake nodes into adversarial text.
\subsubsection{Evaluation Metrics.}
% We use three widely adopted metrics for evaluation. Following \cite{zheng2021graph}, we adopt classification accuracy as the primary evaluation metric. We also report 3-Max and Weighted Accuracy. All results are averaged over five random seeds. 
% More details of the experimental setup are provided in the \textbf{Appendix}.
% Following \cite{zheng2021graph}, we adopt classification accuracy as the primary evaluation metric \textcolor{red}{in the Appendix}. In addition, we report 3-Max, which reflects the average accuracy of the top-3 most robust defenses, and Weighted Accuracy, which assigns different weights to defenses based on their robustness \textcolor{red}{in the Appendix}. Specifically, more robust defenses receive higher weights, as detailed in: 
% $s^{\text{ATK}}_w = \sum_{i=1}^{n} w_i s_i$, $w_i = \frac{1/i^2}{\sum_{j=1}^{n} (1/j^2)}$, $s_i = (S^{\text{DEF}}_{\text{descend}})_i$, 
% where $S^{\text{DEF}}_{\text{descend}}$ is the set of defense scores in a descending order.  
% We evaluate attack performance using three metrics. 
% Following~\cite{zheng2021graph}, we report \textbf{Accuracy} as the primary metric, and further include \textbf{3-Max Accuracy}, which measures the average performance against the top-3 most robust defenses, and \textbf{Weighted Accuracy}, which assigns defense-specific weights based on robustness levels. All results are averaged over five random seeds.  
% We evaluate attack performance using three metrics.  
Following~\cite{zheng2021graph}, we report \textbf{Accuracy} as the primary metric, along with \textbf{3-Max Accuracy}—the average performance against the three most robust defenses—and \textbf{Weighted Accuracy}, which weights results by defense robustness.  
All results are averaged over five random seeds.

\begin{table*}
\small
\centering
\caption{\textbf{GNN-as-Reasoner Scenarios.} Attack performance on three datasets across various target GNNs.
Lower scores indicate stronger attacks.
"Clean" denotes unperturbed graphs; 
\textit{Our-text} and \textit{Our-struct} correspond to text-only and structure-only variants.
Best results are in bold.}
\label{tab:overall-performance}
\resizebox{\linewidth}{!}{%
\begin{tabular}{lclcccccccccc} 
\hline
Dataset & \multicolumn{1}{l}{Emb.} & Models & Clean & STACK & PGD & RND & FLIP & SGAttack & WTGIA & Our-text & Our-struct & \textsc{BadGraph} \\ 
\hline
\multirow{15}{*}{\rotatebox{90}{Cora}} 
 & \multirow{5}{*}{\rotatebox{90}{TF-IDF}} 
 & R-GCN     & {\cellcolor[rgb]{0.898,0.898,0.898}}86.67~ & 82.37~ & 78.07~ & 73.93~ & 73.11~ & 85.41~   & 86.22~ & 84.44~   & 79.11~     & \textbf{70.89~}  \\
 & & GIN       & {\cellcolor[rgb]{0.898,0.898,0.898}}85.93~ & 72.37~ & 78.30~ & 72.59~ & 72.89~ & 81.04~   & 85.70~ & 84.44~   & 76.96~     & \textbf{63.04~}  \\
 & & GraphSAGE & {\cellcolor[rgb]{0.898,0.898,0.898}}84.96~ & 72.30~ & 77.41~ & 74.81~ & 74.74~ & 73.70~   & 83.19~ & 73.93~   & 76.37~     & \textbf{54.67~}  \\
 & & TAGCN     & {\cellcolor[rgb]{0.898,0.898,0.898}}87.33~ & 75.33~ & 80.22~ & 75.04~ & 75.85~ & 87.33~   & 87.56~ & 85.48~   & 83.41~     & \textbf{73.93~}  \\
 & & GCN       & {\cellcolor[rgb]{0.898,0.898,0.898}}85.48~ & 73.78~ & 77.70~ & 74.89~ & 74.44~ & 83.63~   & 83.85~ & 84.15~   & 78.67~     & \textbf{71.33~}  \\ 
\cline{2-13}
 & \multirow{5}{*}{\rotatebox{90}{SBERT}}  
 & R-GCN     & {\cellcolor[rgb]{0.898,0.898,0.898}}87.33~ & 76.37~ & 80.30~ & 76.00~ & 75.93~ & 86.30~   & 85.93~ & 80.30~   & 79.11~     & \textbf{56.67~}  \\
 & & GIN       & {\cellcolor[rgb]{0.898,0.898,0.898}}85.70~ & 73.85~ & 80.15~ & 75.33~ & 75.93~ & 81.04~   & 84.96~ & 77.70~   & 76.00~     & \textbf{47.04~}  \\
 & & GraphSAGE & {\cellcolor[rgb]{0.898,0.898,0.898}}86.00~ & 75.93~ & 79.48~ & 76.00~ & 77.56~ & 72.44~   & 83.33~ & 64.30~   & 70.15~     & \textbf{39.93~}  \\
 & & TAGCN     & {\cellcolor[rgb]{0.898,0.898,0.898}}88.22~ & 78.00~ & 80.81~ & 76.74~ & 75.85~ & 88.74~   & 87.93~ & 79.04~   & 83.26~     & \textbf{51.19~}  \\
 & & GCN       & {\cellcolor[rgb]{0.898,0.898,0.898}}87.11~ & 77.63~ & 80.59~ & 79.04~ & 77.33~ & 86.37~   & 86.30~ & 80.15~   & 78.07~     & \textbf{56.00~}  \\ 
\cline{2-13}
 & \multirow{5}{*}{\rotatebox{90}{TAPE}}   
 & R-GCN     & {\cellcolor[rgb]{0.898,0.898,0.898}}89.63~ & 82.96~ & 84.15~ & 83.33~ & 86.52~ & 87.85~   & -      & 86.30~   & 84.07~     & \textbf{72.89~}  \\
 & & GIN       & {\cellcolor[rgb]{0.898,0.898,0.898}}88.52~ & 80.96~ & 84.22~ & 81.70~ & 84.22~ & 84.22~   & -      & 83.11~   & 82.96~     & \textbf{65.56~}  \\
 & & GraphSAGE & {\cellcolor[rgb]{0.898,0.898,0.898}}87.85~ & 81.26~ & 83.41~ & 82.81~ & 85.19~ & 74.00~   & -      & 82.81~   & 75.63~     & \textbf{48.89~}  \\
 & & TAGCN     & {\cellcolor[rgb]{0.898,0.898,0.898}}88.07~ & 81.19~ & 84.37~ & 83.04~ & 82.52~ & 87.85~   & -      & 84.96~   & 82.67~     & \textbf{69.56~}  \\
 & & GCN       & {\cellcolor[rgb]{0.898,0.898,0.898}}89.70~ & 83.19~ & 83.56~ & 83.19~ & 85.33~ & 87.11~   & -      & 85.93~   & 80.89~     & \textbf{69.93~}  \\ 
\hline
\multirow{15}{*}{\rotatebox{90}{Arxiv}} 
 & \multirow{5}{*}{\rotatebox{90}{TF-IDF}} 
 & R-GCN     & {\cellcolor[rgb]{0.898,0.898,0.898}}67.98~ & 65.78~ & 62.50~ & 61.34~ & 60.08~ & -        & 61.30~ & 67.46~   & 59.42~     & \textbf{56.56~}  \\
 & & GIN       & {\cellcolor[rgb]{0.898,0.898,0.898}}63.14~ & 58.76~ & 58.16~ & 57.82~ & 56.24~ & -        & 48.82~ & 59.20~   & 51.26~     & \textbf{41.16~}  \\
 & & GraphSAGE & {\cellcolor[rgb]{0.898,0.898,0.898}}61.34~ & 56.58~ & 55.86~ & 55.84~ & 54.02~ & -        & 59.44~ & 47.32~   & 52.90~     & \textbf{34.90~}  \\
 & & TAGCN     & {\cellcolor[rgb]{0.898,0.898,0.898}}68.56~ & 63.90~ & 64.32~ & 63.34~ & 62.40~ & -        & 63.24~ & 64.92~   & 65.66~     & \textbf{59.02~}  \\
 & & GCN       & {\cellcolor[rgb]{0.898,0.898,0.898}}65.40~ & 65.58~ & 61.02~ & 60.88~ & 58.57~ & -        & 64.82~ & 64.46~   & 61.32~     & \textbf{58.56~}  \\ 
\cline{2-13}
 & \multirow{5}{*}{\rotatebox{90}{SBERT}}  
 & R-GCN     & {\cellcolor[rgb]{0.898,0.898,0.898}}67.90~ & 67.08~ & 63.48~ & 65.00~ & 62.70~ & -        & 61.22~ & 67.56~   & 60.94~     & \textbf{58.94~}  \\
 & & GIN       & {\cellcolor[rgb]{0.898,0.898,0.898}}69.04~ & 66.78~ & 64.98~ & 65.38~ & 63.40~ & -        & 63.36~ & 66.96~   & 56.82~     & \textbf{44.26~}  \\
 & & GraphSAGE & {\cellcolor[rgb]{0.898,0.898,0.898}}70.00~ & 67.42~ & 66.92~ & 65.66~ & 64.80~ & -        & 64.28~ & 62.14~   & 66.32~     & \textbf{47.58~}  \\
 & & TAGCN     & {\cellcolor[rgb]{0.898,0.898,0.898}}70.30~ & 68.30~ & 66.90~ & 67.32~ & 65.58~ & -        & 69.82~ & 65.88~   & 68.00~     & \textbf{60.28~}  \\
 & & GCN       & {\cellcolor[rgb]{0.898,0.898,0.898}}67.02~ & 68.28~ & 63.80~ & 63.34~ & 62.72~ &          & 67.86~ & 66.58~   & 62.72~     & \textbf{59.30~}  \\ 
\cline{2-13}
 & \multirow{5}{*}{\rotatebox{90}{TAPE}}   
 & R-GCN     & {\cellcolor[rgb]{0.898,0.898,0.898}}70.00~ & 75.82~ & 66.96~ & 69.44~ & 68.18~ & -        & -      & 69.20~   & 66.54~     & \textbf{65.22~}  \\
 & & GIN       & {\cellcolor[rgb]{0.898,0.898,0.898}}77.68~ & 77.06~ & 78.10~ & 78.04~ & 78.48~ & -        & -      & 56.42~   & 63.62~     & \textbf{25.22~}  \\
 & & GraphSAGE & {\cellcolor[rgb]{0.898,0.898,0.898}}80.82~ & 80.44~ & 79.84~ & 79.90~ & 80.12~ & -        & -      & 52.70~   & 80.12~     & \textbf{31.44~}  \\
 & & TAGCN     & {\cellcolor[rgb]{0.898,0.898,0.898}}80.48~ & 81.40~ & 79.40~ & 79.16~ & 79.34~ & -        & -      & 64.42~   & 79.80~     & \textbf{58.62~}  \\
 & & GCN       & {\cellcolor[rgb]{0.898,0.898,0.898}}70.24~ & 77.92~ & 69.36~ & 70.58~ & 69.86~ & -         & -       & 69.22~   & 68.46~     & \textbf{66.14~}  \\ 
\hline
\multirow{15}{*}{\rotatebox{90}{Products}} 
 & \multirow{5}{*}{\rotatebox{90}{TF-IDF}} 
 & R-GCN     & {\cellcolor[rgb]{0.898,0.898,0.898}}86.00~ & 80.70~ & 83.28~ & 81.54~ & 81.58~ & 85.28~   & 82.08~ & 81.84~   & 81.56~     & \textbf{70.84~}  \\
 & & GIN       & {\cellcolor[rgb]{0.898,0.898,0.898}}85.28~ & 79.78~ & 82.58~ & 81.62~ & 81.16~ & 81.62~   & 68.42~ & 75.90~   & 75.18~     & \textbf{53.30~}  \\
 & & GraphSAGE & {\cellcolor[rgb]{0.898,0.898,0.898}}83.94~ & 78.18~ & 81.42~ & 80.78~ & 79.36~ & 80.02~   & 80.90~ & 70.14~   & 71.02~     & \textbf{50.06~}  \\
 & & TAGCN     & {\cellcolor[rgb]{0.898,0.898,0.898}}85.94~ & 78.72~ & 82.90~ & 82.42~ & 79.50~ & 85.32~   & 83.40~ & 83.16~   & 82.68~     & \textbf{74.14~}  \\
 & & GCN       & {\cellcolor[rgb]{0.898,0.898,0.898}}86.48~ & 82.12~ & 84.28~ & 82.28~ & 82.98~ & 85.38~   & 83.92~ & 81.20~   & 81.90~     & \textbf{69.04~}  \\ 
\cline{2-13}
 & \multirow{5}{*}{\rotatebox{90}{SBERT}}  
 & R-GCN     & {\cellcolor[rgb]{0.898,0.898,0.898}}87.38~ & 82.98~ & 86.34~ & 84.48~ & 84.16~ & 86.92~   & 80.84~ & 86.02~   & 84.66~     & \textbf{79.06~}  \\
 & & GIN       & {\cellcolor[rgb]{0.898,0.898,0.898}}86.08~ & 81.54~ & 84.28~ & 83.06~ & 82.86~ & 84.02~   & 83.60~ & 84.24~   & 80.40~     & \textbf{71.18~}  \\
 & & GraphSAGE & {\cellcolor[rgb]{0.898,0.898,0.898}}87.08~ & 82.76~ & 85.74~ & 84.64~ & 84.14~ & 83.46~   & 81.38~ & 80.72~   & 73.48~     & \textbf{58.60~}  \\
 & & TAGCN     & {\cellcolor[rgb]{0.898,0.898,0.898}}88.08~ & 80.40~ & 86.42~ & 84.82~ & 82.30~ & 87.60~   & 86.34~ & 86.72~   & 85.26~     & \textbf{80.28~}  \\
 & & GCN       & {\cellcolor[rgb]{0.898,0.898,0.898}}87.26~ & 83.66~ & 86.20~ & 84.18~ & 84.26~ & 86.86~   & 85.66~ & 86.36~   & 84.04~     & \textbf{77.58~}  \\ 
\cline{2-13}
 & \multirow{5}{*}{\rotatebox{90}{TAPE}}   
 & R-GCN     & {\cellcolor[rgb]{0.898,0.898,0.898}}89.54~ & 89.58~ & 89.22~ & 88.86~ & 89.96~ & 89.38~   & -      & 87.42~   & 88.56~     & \textbf{80.52~}  \\
 & & GIN       & {\cellcolor[rgb]{0.898,0.898,0.898}}88.64~ & 88.80~ & 88.58~ & 88.02~ & 88.88~ & 87.48~   & -      & 82.26~   & 86.82~     & \textbf{69.12~}  \\
 & & GraphSAGE & {\cellcolor[rgb]{0.898,0.898,0.898}}90.22~ & 90.70~ & 90.10~ & 89.86~ & 90.48~ & 90.16~   & -      & 81.06~   & 89.58~     & \textbf{66.28~}  \\
 & & TAGCN     & {\cellcolor[rgb]{0.898,0.898,0.898}}89.08~ & 88.74~ & 88.58~ & 88.30~ & 88.40~ & 89.06~   & -      & 83.84~   & 88.62~     & \textbf{79.98~}  \\
 & & GCN       & {\cellcolor[rgb]{0.898,0.898,0.898}}89.44~ & 89.80~ & 88.96~ & 88.96~ & 89.72~ & 89.32~   & -       & 86.44~   & 88.56~     & \textbf{75.74~}  \\
\hline
\end{tabular}
}
\end{table*}

\subsection{Overall Performance}
\subsubsection{Attack Success Across Backbone Models in GNN-as-Reasoner Scenarios.}
\textsc{BadGraph} framework is highly flexible and can be universally applied to attack any type of target TAG backbone. We first evaluate the performance of \textsc{BadGraph} in a black-box setting across various models in GNN-as-Reasoner scenarios. To this end, we explore three node feature encoding methods with different levels of expressiveness: TF-IDF, SBERT, and TAPE. The results are shown in Table \ref{tab:overall-performance}, we have the following observations:
(1) Compared to existing attack baselines, \textbf{ \textsc{BadGraph} achieves state-of-the-art (SOTA) performance across all target models and datasets.}
These findings further highlight the effectiveness and universality of \textsc{BadGraph} in attacking TAG models. Notably, our method results in an accuracy drop of nearly 40\% on Cora, 52\% on OGBN-Arxiv, and around 30\% on OGBN-Products. 
This performance drop is mainly due to the powerful adversarial perturbations generated by our LLM, which leverage rich graph-related knowledge to pinpoint and exploit the most vulnerable structures and texts.
(2) Surprisingly, \textbf{we find that models with richer node features tend to exhibit greater robustness.} While traditional structure-based attacks are generally effective against shallow features like TF-IDF, they have minimal impact on models leveraging more expressive textual representations such as SBERT and TAPE, on OGBN-Arxiv, some attacks even lead to accuracy improvements. This is because these models effectively capture the semantic information of TAGs, enabling accurate predictions even when the structural information is compromised.  (3) \textbf{Single-modality attacks, whether structural or textual, are ineffective}—our text-only and structure-only variants confirm this through significantly reduced performance, consistent with our cross-modal shortcut theory.

\begin{table*}
\small
\centering
\caption{\textbf{LLM-as-Reasoner Scenarios.} Attack performance across two LLM backbones on three TAG datasets.}
\label{tab:overall-performance-llm}
% \resizebox{\linewidth}{!}{%
\begin{tabular}{llcccccccccc} 
\hline
Dataset                   & Models     & Clean  & STACK  & PGD    & RND    & FLIP   & SGAttack & WTGIA  & \begin{tabular}[c]{@{}c@{}}Our-text\end{tabular} & \begin{tabular}[c]{@{}c@{}}Our-struct\end{tabular} & \textsc{BadGraph}  \\ 
\hline
\multirow{2}{*}{Cora}     & DeepSeek-V3   & \cellcolor{gray!20}71.85~ & 68.14~ & 72.96~ & 65.92~ & 64.07~ & 67.03~   & 63.70~ & 25.55~                                                  & 67.03~                                                    & \textbf{17.03}~    \\
                          & Mistral-7B & \cellcolor{gray!20}43.70~ & 42.96~ & 53.33~ & 42.59~ & 47.03~ & 37.40~   & 21.48~ & 21.11~                                                  & 38.51~                                                    & \textbf{16.29}~    \\ 
\hline
\multirow{2}{*}{Arxiv}    & DeepSeek-V3   & \cellcolor{gray!20}64.60~ & 60.80~ & 60.60~ & 61.20~ & 58.70~ & -        & 60.10~ & 44.30~                                                  & 55.55~                                                    & \textbf{22.90}~    \\
                          & Mistral-7B & \cellcolor{gray!20}9.30~  & 7.70~  & 8.90~  & 9.20~  & 9.00~  & -        & 7.60~  & 4.90~                                                   & 4.70~                                                     & \textbf{2.90}~     \\ 
\hline
\multirow{2}{*}{Products} & DeepSeek-V3   & \cellcolor{gray!20}77.10~ & 75.90~ & 75.00~ & 76.50~ & 75.60~ & 75.80~   & 74.10~ & 60.20~                                                  & 75.20~                                                    & \textbf{55.80}~    \\
                          & Mistral-7B & \cellcolor{gray!20}12.20~ & 9.50~  & 10.00~ & 9.70~  & 8.30~  & 12.20~   & 9.10~  & 8.60~                                                   & 12.80~                                                    & \textbf{7.80}~     \\
\hline
\end{tabular}
% }
\end{table*}

\subsubsection{Attack Success Across Backbone Models in LLM-as-Reasoner Scenarios.}
To develop a more generalizable model, we evaluate the performance of \textsc{BadGraph} in LLM-as-Reasoner scenarios, where the graph structure and textual attributes are converted into prompts understandable by the LLM, and the LLM makes predictions based on its reasoning capabilities. The results are presented in Table \ref{tab:overall-performance-llm}. Additionally, we observed several interesting findings: (1) \textbf{\textsc{BadGraph} consistently demonstrates strong destructive power against two target LLMs across all datasets}, indicating its effectiveness in undermining the LLMs’ ability to comprehend graph patterns. Specifically, on the Cora dataset under the same perturbation budget, \textsc{BadGraph} reduces DeepSeek's prediction accuracy by 76.3\%, whereas the best-performing baseline, WTGIA, results in only an 11.3\% drop. This significant degradation stems from \textsc{BadGraph}’s ability to uncover the underlying principles of how LLMs interpret graph structures—and to precisely disrupt them.
(2) \textbf{LLM-as-Reasoner models tend to be more vulnerable to text-based attacks while exhibiting greater resilience to structural perturbations.} For instance, a text-only attack using \textsc{BadGraph} reduces DeepSeek's prediction accuracy by 64\%, whereas a structure-only attack leads to just a 6.3\% drop. This disparity arises because LLMs primarily rely on textual information to infer graph patterns—a stark contrast to the GNN-as-Reasoner setting, where models place greater emphasis on structural semantics. Overall, irrespective of the underlying backbone, jointly attacking both text and structure consistently achieves the most impactful results.

\begin{figure}[htbp]
  \centering

  \begin{subfigure}[t]{0.21\textwidth}
    \includegraphics[width=\textwidth]{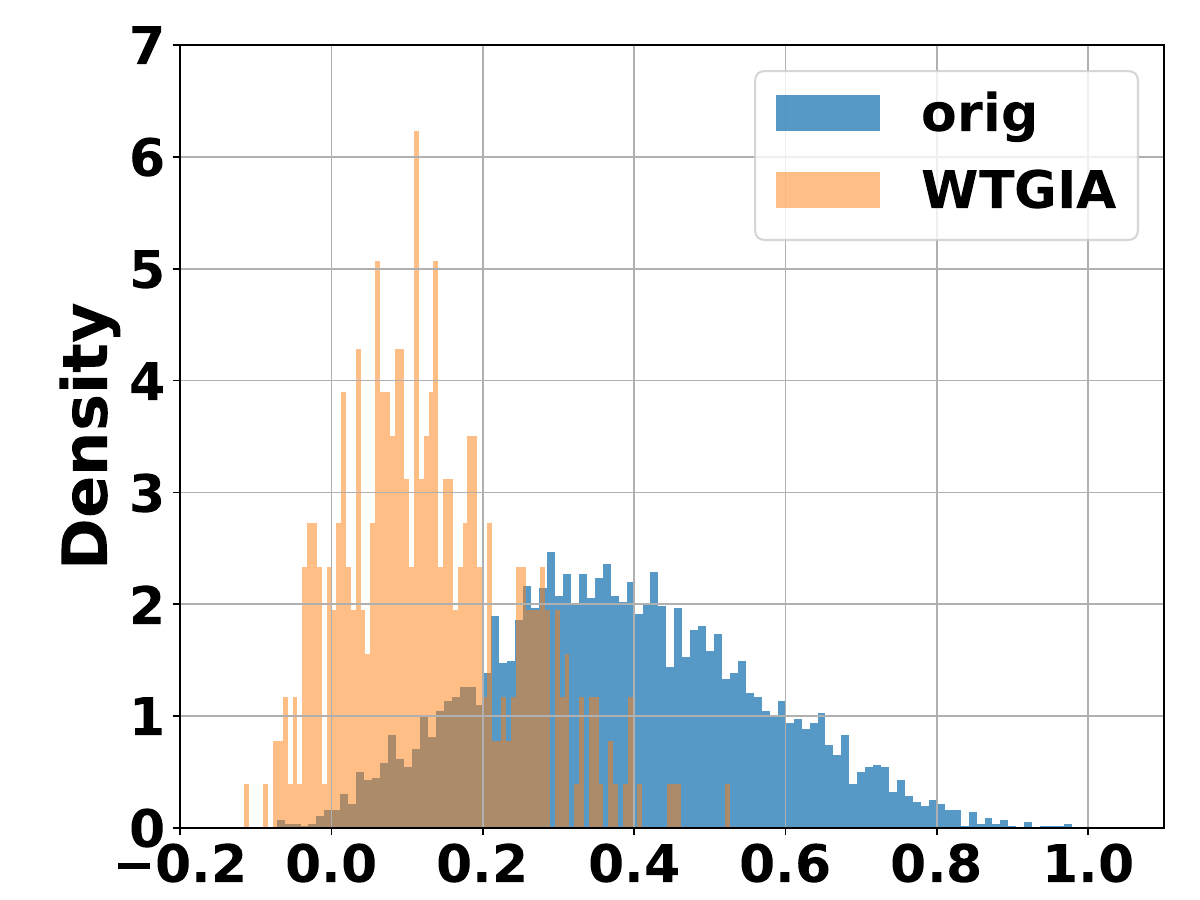}
    \caption{WTGIA}
  \end{subfigure}
  \hfill
  \begin{subfigure}[t]{0.21\textwidth}
    \includegraphics[width=\textwidth]{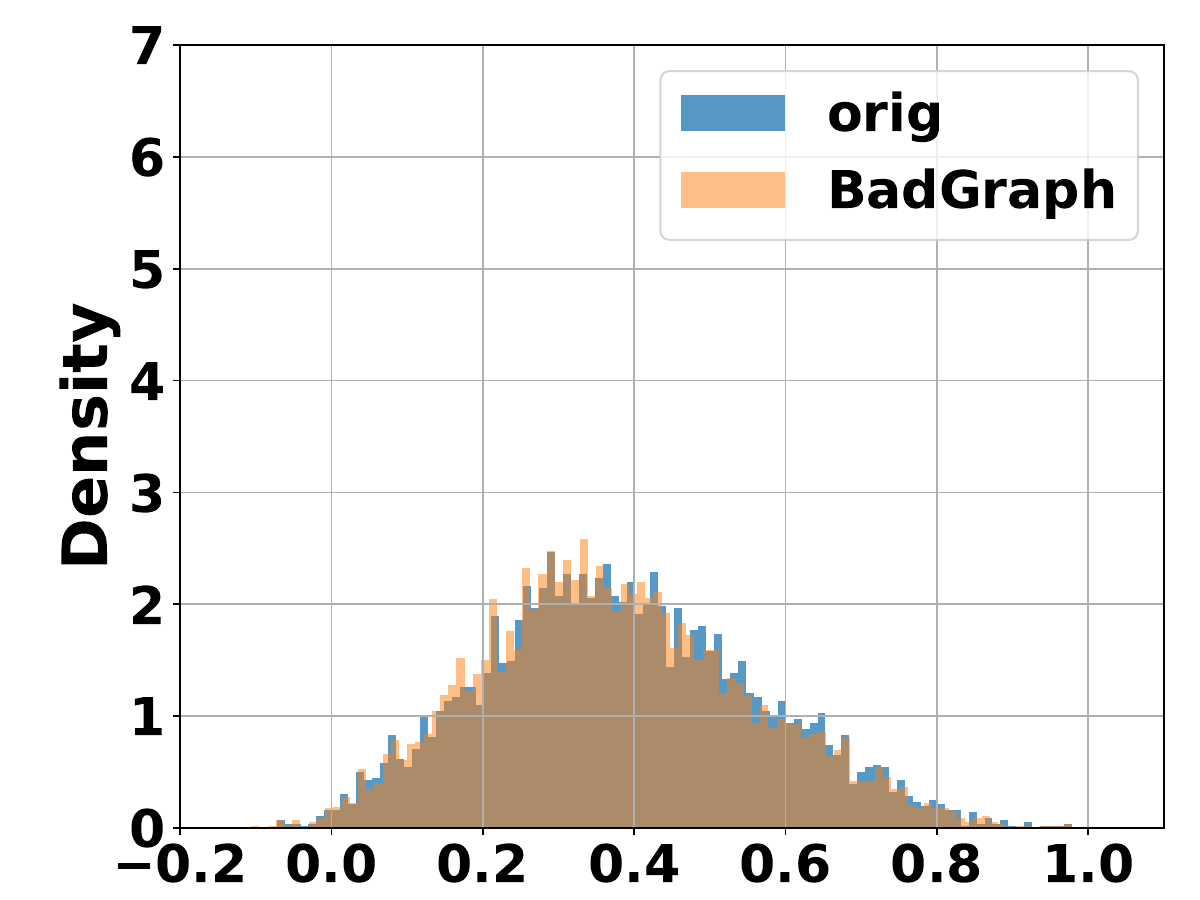}
    \caption{\textsc{BadGraph}}
  \end{subfigure}

  \caption{Edge-level homophily shift on the Cora dataset.}
  \label{fig:homo}
\end{figure}

\begin{table}
\small
\centering
\caption{Accuracy and homophily changes of deepSeek-as-Reasoner on Arxiv under varying perturbation ratios.}
\label{tab:Homophily_Changes}
% \resizebox{\linewidth}{!}{%
\begin{tabular}{ccccc} 
\hline
$\#$Nodes & \begin{tabular}[c]{@{}c@{}}Perturb. Ratio\end{tabular} & Acc    & \begin{tabular}[c]{@{}c@{}} Avg. Homo.\\~(edge)\end{tabular} & \begin{tabular}[c]{@{}c@{}}Avg. Homo. \\~(node)\end{tabular}  \\ 
\hline
0           & 0                                                              & 64.6\% & 0.8184                                                           & 0.4257                                                             \\
500         & 0.09\%                                                         & 23.4\% & 0.8183                                                           & 0.4292                                                             \\
1000        & 0.17\%                                                         & 22.9\% & 0.8182                                                           & 0.4317                                                             \\
1500        & 0.26\%                                                         & 22.6\% & 0.8181                                                           & 0.4316                                                             \\
\hline
\end{tabular}
% }
\end{table}

% \begin{table}
% \centering
% \caption{Efficiency analysis of \textsc{BadGraph}.}
% \label{efficiency}
% % \small % 控制表格字体大小
% \begin{tabular}{cccc} 
% \toprule
% Dataset & Time (s) & Mem. (MB) & Cost \\ 
% \midrule
% Cora           & 34               & 820.5             &     \$0.0009\\
% Arxiv     & 37.3             & 1011.73           &     \$0.0009\\
% Products  & 33.2             & 1037.1            &     \$0.0008\\
% \bottomrule
% \end{tabular}
% \end{table}

\subsection{Merits of \textsc{BadGraph}}
\subsubsection{Stealthy, Localized Attacks with Homophily Stability}

\textbf{Observation 1.} \textbf{\textsc{BadGraph} maintains global homophily stability. } 
We evaluate the imperceptibility of \textsc{BadGraph} using edge-centric homophily metrics~\cite{chen2022understandingimprovinggraphinjection}.  
As shown in Fig.~\ref{fig:homo}, WTGIA causes pronounced shifts in homophily distributions, making it easily detectable by homophily-based defenses.  
In contrast, \textsc{BadGraph} introduces only subtle and localized distributional changes, confirming its stealthy and unnoticeable nature.  
This stability stems from semantically aligned, anchor-guided perturbations that jointly adjust topology and text while preserving the global structural context.
\textbf{Observation 2.}\label{sec:empirical_stealthiness} \textbf{\textsc{BadGraph} achieves high attack effectiveness without disrupting overall graph properties.}  
As presented in Table~\ref{tab:Homophily_Changes}, even when the perturbation budget increases (500~$\rightarrow$~1500 edges), node- and edge-level homophily remain nearly unchanged, whereas target node accuracy drops sharply.  
This demonstrates that \textsc{BadGraph} performs tightly localized “edge rewriting,” where each deletion is paired with an addition, maintaining structural balance while creating semantic shortcuts.  
Although a small fraction of semantically important edges may be modified (\(\leq 0.26\%\)), these edits are context-aware and minimally invasive, preserving the global homophily distribution. 

\subsubsection{Affordable Inference Cost and High Destructiveness.}

\begin{figure}[t]
	{
		\begin{minipage}[t]{0.48\linewidth}
			\centering
			\includegraphics[width=1\textwidth]{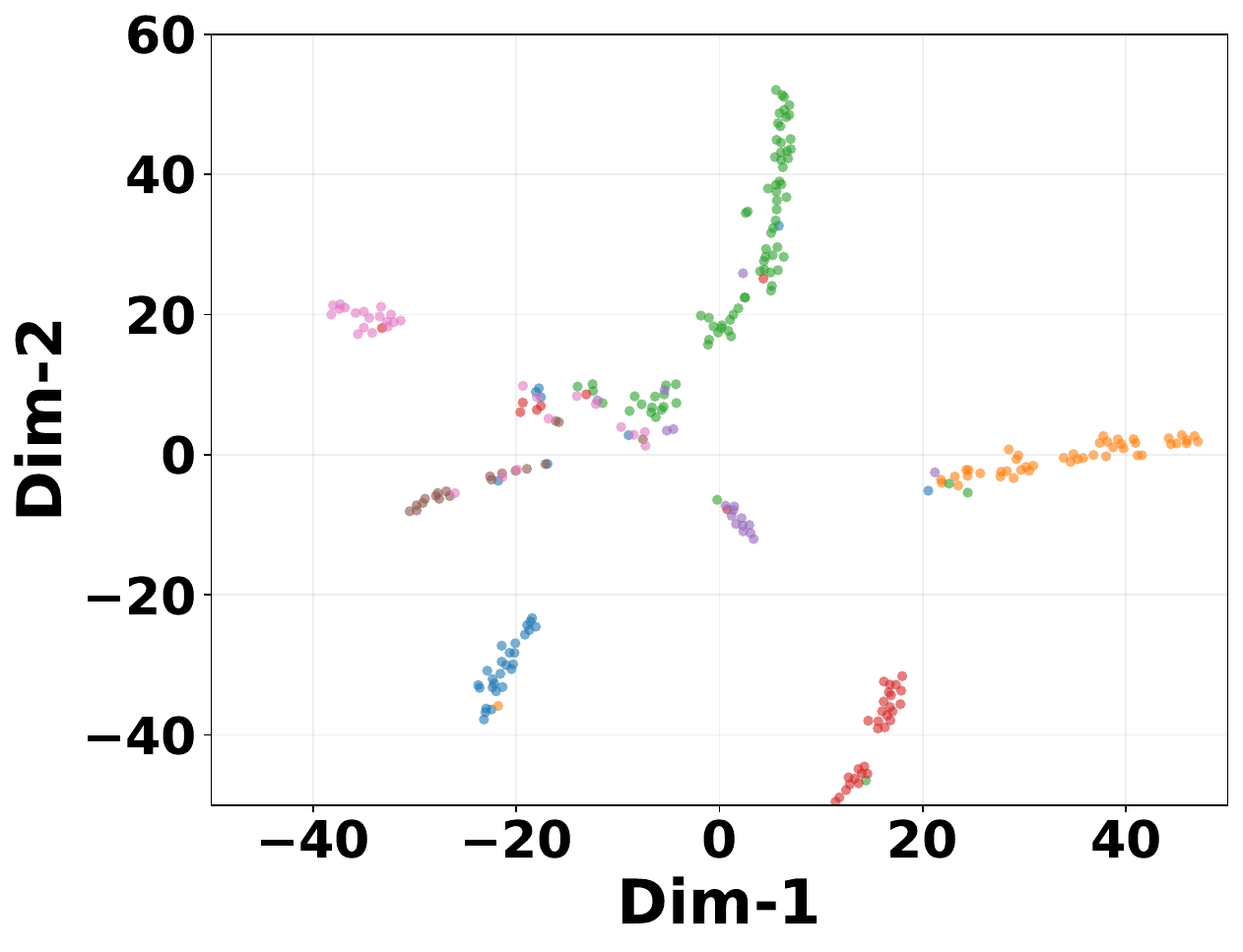}
			% \vspace{-0.1in}
			\subcaption{WTGIA}
		\end{minipage}
		\begin{minipage}[t]{0.48\linewidth}
			\centering
			\includegraphics[width=1\textwidth]{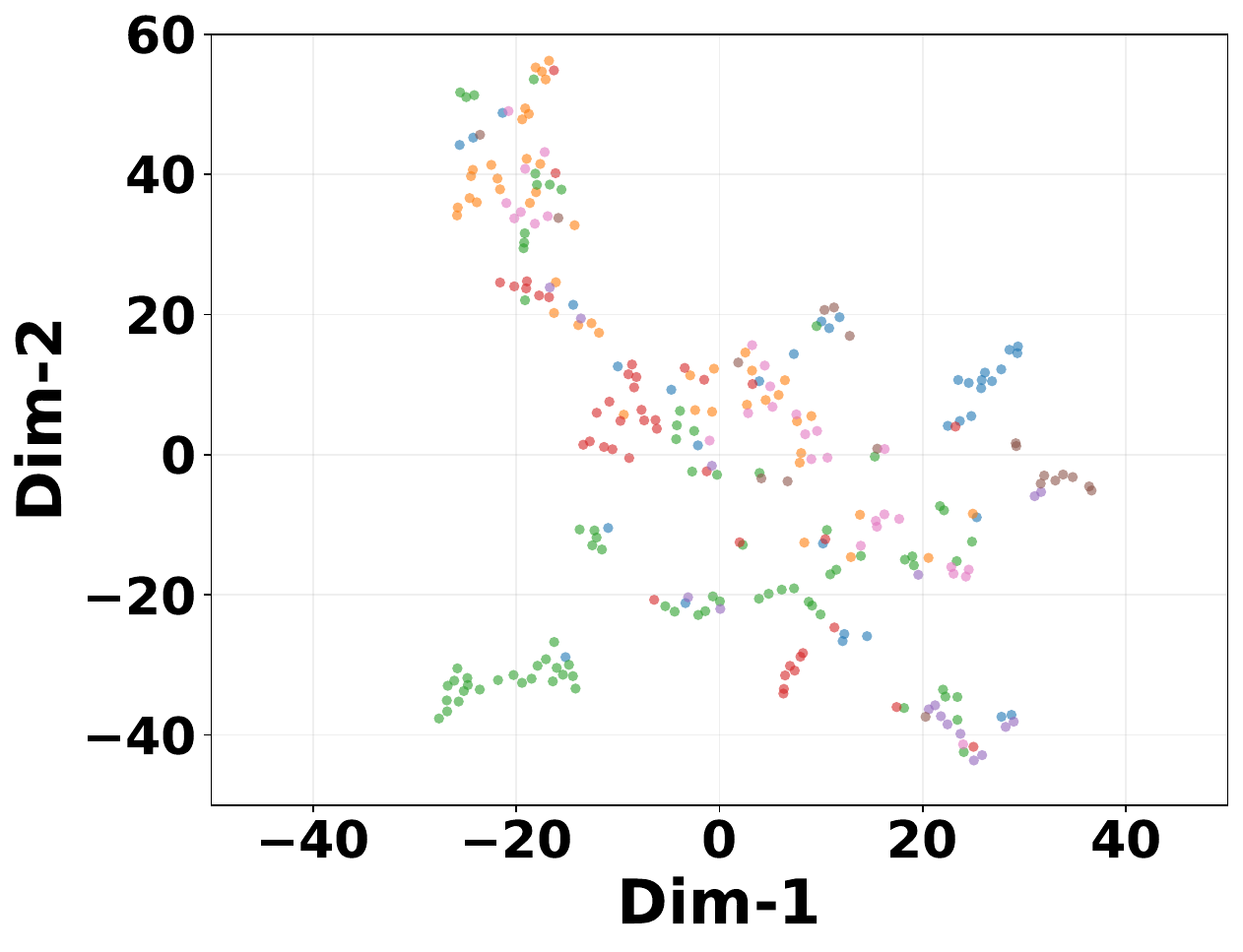}
			% \vspace{-0.1in}
			\subcaption{\textsc{BadGraph}}
		\end{minipage}
	}
  \caption{T-SNE visualization illustrating embedding shifts caused by the attack on the Cora dataset.}
  \label{fig:t-sne}
\end{figure}

\textbf{Observation 1.} \textbf{\textsc{BadGraph} achieves efficient and low-cost attack inference. } 
The total computational cost includes (1) the influencer retrieval module and (2) LLM API queries.  
The retriever has a training complexity of $\mathcal{O}(2KNd^2)$, where $K$ is the number of attention heads, $N$ the number of nodes, and $d$ the feature dimension.  
Unlike conventional LLM-based edge prediction approaches that require $\mathcal{O}(N)$ evaluations per target node, our  retrieval-then-reasoning selection reduces the cost to $\mathcal{O}(1)$.  
Moreover, \textsc{BadGraph} executes entirely through API calls on CPU, with an average query cost of approximately~\textdollar0.0009 per node—demonstrating its practicality and scalability for real-world black-box attack scenarios.  
\textbf{Observation 2.} \textbf{\textsc{BadGraph} delivers strong representational disruption despite its lightweight cost.  }
As illustrated in Fig.~\ref{fig:t-sne}, the T-SNE visualization reveals that \textsc{BadGraph} induces significantly greater boundary distortion and class entanglement than WTGIA, implying a deeper perturbation of the latent representation space.  
This indicates that even under minimal resource consumption, \textsc{BadGraph} can generate substantial embedding shifts—achieving a rare balance between \emph{efficiency} and \emph{destructiveness}.

\subsubsection{Robustness to Prompt Design and LLM Variants}

\textbf{Observation 1.} \textbf{\textsc{BadGraph} remains robust to prompt variations, while cross-modal shortcut alignment serves as the key attack guarantee.}
As shown in Table~\ref{tab:different_prompt}, we evaluate multiple prompt phrasings for generating perturbations (e.g., rewording “Summarize why [node] is adjacent to…” as “Explain the likely reason for its connection to…”).  
\textsc{BadGraph} maintains stable attack performance, with accuracy fluctuations within 3.7\%, indicating low sensitivity to surface-level linguistic changes.  
However, replacing the shared influencer node between textual and structural attacks leads to a sharp performance drop (up to 22.66\%), demonstrating that the attack’s effectiveness stems from \emph{semantic alignment across modalities} rather than prompt wording itself.
\textbf{Observation 2.} \textbf{\textsc{BadGraph} generalizes across different attacker LLMs.}  
We further test \textsc{BadGraph} with diverse open-source LLMs (e.g., Qwen-Plus,  LLaMA-4-17B), as reported in Table~\ref{tab:different_llm_attacker}.  
Across all models, \textsc{BadGraph} consistently achieves strong attack effectiveness, while larger models (e.g., Qwen-Plus) produce slightly more precise and context-aware perturbations.  
These results confirm that \textsc{BadGraph} is requiring no model-specific tuning, and can flexibly leverage the reasoning capability of emerging LLMs—underscoring its scalability and plug-and-play design for future LLM-powered TAG attacks.

\begin{table}
\centering
\caption{Robustness of \textsc{BadGraph} under prompt variations on Cora.
“Rephrasing” rewrites the prompt, and “Anchor Mis.” uses inconsistent influencers across modalities.}
\label{tab:different_prompt}
\resizebox{\linewidth}{!}{%
\begin{tabular}{lccccc} 
\hline
Method                                                              & R-GCN          & GCN            & GIN            & GraphSAGE      & TAGCN           \\ 
\hline
Clean                                                               & 87.33          & 87.11          & 85.70          & 86.00          & 88.22           \\
\begin{tabular}[c]{@{}l@{}}Anchor Mis.\end{tabular} & 71.63          & 63.93          & 71.04          & 62.59          & 77.11           \\
\begin{tabular}[c]{@{}l@{}}Rephrasing\end{tabular}      & 58.15          & 58.74          & 50.74          & \textbf{39.04} & 54.59           \\
\textsc{BadGraph}                                                            & \textbf{56.67} & \textbf{56.00} & \textbf{47.04} & 39.93          & \textbf{51.19}  \\
\hline
\end{tabular}
}
\end{table}

% \vspace{-20pt}

\begin{table}
\centering
\caption{\textbf{Robustness of \textsc{BadGraph} under Different Attacker LLMs on the Cora Dataset.}
Attacker models: LLaMA-4-17B, Qwen-Plus, and DeepSeek-V3.}
\label{tab:different_llm_attacker}
\resizebox{\linewidth}{!}{%
\begin{tabular}{lcccccc} 
\hline
Method                                                            & GIN            & GraphSAGE      & SGCN           & TAGCN          & Qwen           & DeepSeek        \\ 
\hline
Clean                                                             & 85.70          & 86.00          & 86.07          & 88.22          & 68.88          & 71.85           \\
\begin{tabular}[c]{@{}l@{}}Ours-LLaMA\end{tabular}  & 48.81          & \textbf{39.85} & 58.52          & 52.44          & \textbf{15.92} & 17.03           \\
\begin{tabular}[c]{@{}l@{}}Ours-Qwen\end{tabular}    & 48.70          & \textbf{39.56} & 55.78          & 55.19          & 17.03          & 19.25           \\
\begin{tabular}[c]{@{}l@{}}Ours-DeepSeek\end{tabular} & \textbf{47.04} & 39.93          & \textbf{55.63} & \textbf{51.19} & 16.29          & \textbf{16.29}  \\
\hline
\end{tabular}
}
\end{table}

\subsubsection{Potential for Interpretable Attack Patterns.}
\label{sec:potial_for_interpretable_atk_pattern}
To demonstrate the reasonableness and interpretability of the attacks generated by \textsc{BadGraph}, we present a case study using DeepSeek to illustrate the reasoning process behind structural attacks. Instead of directly crafting adversarial examples, we reformulate the attack as a harmful prompt generation task, leveraging the reasoning capabilities of LLMs to identify structural patterns and generate effective perturbations. Given a target node (e.g., \textit{A Non-linear Information Maximisation Algorithm that Performs Blind Separation}), \textbf{the LLM produces not only potential perturbations but also natural language justifications for each modification}, revealing the semantic and structural rationale behind the attack.

% \section{Case Study}
% \label{appendix:inter_atk_pattern}
\begin{tcolorbox}[
  colback=gray!5!white,
  colframe=gray!40!black,
  boxrule=0.2pt,
  left=2pt,right=2pt,top=2pt,bottom=2pt,
  arc=2pt,
  breakable,
]
\textcolor{blue}{Step 1: Analysis of the Target Node and Its Neighboring Set}

\textbf{Summary of Adjacency and Correlation:}

The neighboring nodes are adjacent to the target node due to shared themes in unsupervised learning, information maximization, and blind source separation. The most prominent factors guiding their strong correlation are:  

- Information Maximization: The target node and the first neighboring node explicitly focus on information-theoretic approaches for blind separation.  

- Unsupervised Learning: All neighboring nodes address unsupervised methods (competitive learning, feature extraction, or classification).  

- Blind Separation/Deconvolution: The first neighboring node directly extends the target node's core problem (blind separation).

\textcolor{blue}{Step 2: Most Relevant Node in the Neighboring Set}

\textbf{[Node]} \textit{"An information-maximisation approach to blind separation and blind deconvolution"  }

\textbf{Reason: }This node is the most relevant because it:  

1. Shares the identical problem domain (blind separation/deconvolution).  

2. Builds on the same theoretical foundation (information maximization).  

3. Likely cites the target node as prior work or a direct extension.

\textcolor{blue}{Step 3: Exclusion of Least Relevant Node from Candidate List}

\textbf{[Node, Reason for irrelevance]} 

\textit{"Type Inheritance in Strongly Typed Genetic Programming"}  

\textbf{Reason:} This paper discusses type systems in genetic programming, a niche topic far removed from the target node's focus on information maximization and blind separation. The other candidates at least tangentially relate to optimization (closer to machine learning), but this one is purely about programming language design for GAs.  

\end{tcolorbox}

% \begin{figure*}[htbp]
%     \centering
%     \begin{minipage}[b]{0.30\linewidth}
%         \centering
%         \includegraphics[width=\linewidth]{figs/low-degree/sbert_plot.pdf}

%     \end{minipage}
%     \hfill
%     \begin{minipage}[b]{0.30\linewidth}
%         \centering
%         \includegraphics[width=\linewidth]{figs/low-degree/sbert_plot.pdf}
%         % \caption{Fig 2}
%     \end{minipage}
%     \hfill
%     \begin{minipage}[b]{0.36\linewidth}
    
%         \centering
%         \small
%         \setlength{\tabcolsep}{3pt}
%         \resizebox{\linewidth}{!}{
%         \begin{tabular}{lccc}
%         \hline
%         Dataset   & Time (s) & Memory (MB) & Cost \\
%         \midrule
%         Cora        & 34       & 820.5       & \$0.0009 \\
%         Arxiv       & 37.3     & 1011.73     & \$0.0009 \\
%         Products    & 33.2     & 1037.1      & \$0.0008 \\
%         \hline
%         \end{tabular}
%         }
%         \captionof{table}{Efficiency analysis.}
%     \end{minipage}
%     \caption{Prediction using different models and visualizations.}
% \end{figure*}

\section{Conclusion}
In this paper, we propose BadGraph, a novel method that enables LLMs to generate universal perturbations targeting both node topology and textual semantics in TAGs, all in a black-box manner. We design a target influencer retrieval module that identifies candidate influencer nodes, providing attack evidence and facilitating the construction of cross-modal adversarial shortcuts.
The resulting adversarial graph is backbone-agnostic and transfers effectively across diverse graph learning models.
Guided by a homophily-preserving bound and a cross-modal shortcut theory, we explain why \textsc{BadGraph} achieves both high attack efficacy and stealthiness. 
Extensive experiments show that it substantially degrades GNN- and LLM-based TAG backbones while producing interpretable, hard-to-detect, and cost-efficient attacks. 
As the first LLM-driven TAG attacker, this work lays a simple yet effective foundation for future studies, including autonomous LLM agents for adaptive attacks.
% Experimental results show that BadGraph significantly degrades the performance of multiple TAG backbones. 
% Furthermore, additional analyses reveal that the generated attacks are interpretable, stealthy, and cost-effective, underscoring the practical threat posed by LLM-driven adversarial examples in graph learning.

\begin{acks}
This work was supported in part by the Zhejiang Provincial Natural Science Foundation under Grant No. LQN26F020049, in part by the Zhejiang Province Key R\&D Program Project under Grant No. 2025C01023, and in part by the National Natural Science Foundation of China under Grant Nos. 62372146, 62322203, and 62172052.
\end{acks}
\clearpage
\bibliographystyle{ACM-Reference-Format}
\bibliography{references}
\appendix
\section{Details of Experimental Settings}
\subsection{Datasets}
\label{appendix:dataset}
We conduct extensive experiments on three TAG datasets: Cora \cite{mccallum2000automating}, OGBN-Arxiv, and OGBN-Products \cite{hu2020open}. The dataset statistics are provided in Table \ref{datasets}, with the largest dataset containing 1,166,243 edges. We follow the data splitting scheme from \cite{zheng2021graph}, using the full splits for each dataset to train the target model. For the attack settings, we use the medium split as target nodes in Cora. In OGBN-Arxiv, we randomly select 1000 nodes from the medium split as target nodes, and in OGBN-Products, we randomly select 1000 nodes from the hard split. 
To ensure consistency and simplicity, we use minimal textual descriptions for each node in the TAG. Specifically, for the Cora and OGBN-Arxiv datasets, we extract only the paper titles as input text. For the OGBN-Products dataset, we use the product name as the textual input; if the name is unavailable, we substitute it with the product description.
% For the Cora and OGBN-Arxiv datasets, we formatted the raw texts by extracting the title. For the OGBN-Products dataset, we formatted the raw texts using the product name; if the name is missing, we used the product description instead. 
Each experiment is conducted five times, and the average performance and standard deviation are reported. The sources of the datasets are detailed in the footnote\footnote{https://github.com/XiaoxinHe/TAPE}.
% \begin{itemize}
% \item \textbf{Cora, OGBN-Arxiv, OGBN-Products}: \url{https://github.com/XiaoxinHe/TAPE}
% \end{itemize}

\begin{table}[!h]
\centering
\caption{Overview of dataset statistics.}
\label{datasets}
\begin{tabular}{lccc}
\toprule
Dataset & \#Nodes & \#Edges & \#Classes \\
\midrule
Cora & 2,708 & 5,429 & 7 \\
Arxiv & 169,343 & 1,166,243 & 40 \\
Products (subset) & 54,025 & 74,420 & 47 \\
\bottomrule
\end{tabular}
\end{table}

% \subsubsection{Details of Backbones}
% We use DeepRobust and GRB, \textcolor{red}{an} adversarial attack repository, to implement all the attack methods.

\subsection{Target Backbones}
\label{appendix:target_backbone}
Mistral-7B~\cite{jiang2024mixtral} and DeepSeek~\cite{lu2024deepseek} are employed in a zero-shot manner to perform node classification based on textual descriptions and their 2-hop neighborhoods. 

\begin{itemize}[leftmargin=*]
    \item \textbf{GCN} \cite{kipf2016semi}: GCN is a widely used class of graph neural networks grounded in spectral graph theory.
    \item \textbf{GIN} \cite{xu2018powerful}: GIN maximizes the expressive power of GNNs by using injective aggregation functions to distinguish different graph structures.
    \item \textbf{GraphSAGE} \cite{hamilton2017inductive}: GraphSAGE samples a fixed number of neighbors and aggregates their features through learnable functions, enabling inductive learning on large graphs.
    \item \textbf{TAGCN} \cite{du2017topology}: TAGCN leverages fixed-size learnable filters based on powers of the graph Laplacian to capture multi-hop neighborhood information efficiently in a single convolutional layer.
    \item \textbf{SGCN} \cite{wu2019simplifying}: SGCN  simplifies GCN  by merging multiple layers of linear transformations and nonlinear activations into a single neighborhood aggregation step.
    \item \textbf{R-GCN} \cite{zhu2019robust}: R-GCN represents nodes as Gaussian distributions to resist adversarial noise and uses attention to downweight high-variance nodes.
    \item \textbf{RUNG} \cite{hou2024robust}: RUNG employs edge reweighting to prune suspicious edges, thereby enhancing the model's robustness against adversarial attacks.
    % \item \textbf{LLM4RGNN} \cite{zhang2024can}: LLM4RGNN leverages LLMs' reasoning capabilities to purify graph against topology attacks.  
    \item \textbf{DeepSeek} \cite{lu2024deepseek}: DeepSeek adopts a 671B-parameter MoE architecture with MLA and multi-token prediction, demonstrating strong performance on multilingual tasks.
    \item \textbf{Mistral} \cite{jiang2024mixtral}: Mistral is an open-source LLM that adopts Grouped-Query Attention and Sliding Window Attention architectures, offering excellent inference efficiency.
\end{itemize}

\subsection{Baselines}
\label{appendix:baseline}
\begin{itemize}[leftmargin=*]
    \item \textbf{RND} \cite{zugner2018adversarial}: RND is a random attack strategy that only modifies the structure of the graph.
    \item \textbf{FLIP} \cite{bojchevski2019adversarial}: FLIP is a deterministic approach that first ranks all nodes in ascending order according to their degrees, then flips their edges from the lower degree nodes to higher degree nodes.
    \item \textbf{STACK} \cite{xu2012query}: STACK leverages a universal graph filter to unify different graph learning models, using this approximation to perform optimization-based attacks.
    \item \textbf{PGD} \cite{madry2018towards}: PGD works by adjusting inputs in multiple iterative steps towards the direction that increases model loss, thereby generating adversarial samples.
    \item \textbf{NETTACK} \cite{zugner2018adversarial}: NETTACK proposes an efficient algorithm to address the discrete nature of graph data while perturbing both the graph structure and node features.
    \item \textbf{SGAttack} \cite{li2023adversarial}: SGAttack is an efficient, simplified gradient-based attack that performs multi-stage perturbations on target nodes using only a small subgraph. 
    \item \textbf{WTGIA} \cite{lei2024intruding}: WTGIA is a text-level injection attack, utilizing LLMs to convert the embedding information of the fake nodes into adversarial text.
\end{itemize}
\begin{table*}[t]
\centering
\caption{Attack Results Under RUNG Defense on the Cora Dataset.}
\label{tab:attack_rung}

\begin{tabular}{lccccccccc} 
\hline
                                                 & Clean & STACK  & PGD    & RND    & FLIP   & NETTACK & SGAttack & WTGIA  & BadGraph         \\ 
\hline
\begin{tabular}[c]{@{}l@{}}TF-IDF\\\end{tabular} & \cellcolor{gray!20}86.81 & 79.56~ & 82.00~ & 80.89~ & 78.00~ & 86.07   & 87.56~   & 87.26~ & \textbf{62.44~}  \\
SBERT                                            & \cellcolor{gray!20}85.45 & 78.22~ & 81.11  & 78.07~ & 78.15~ & 83.56~  & 86.00~   & 85.11~ & \textbf{35.11~}  \\
\hline
\end{tabular}
\end{table*}

\begin{table}
\centering
\caption{Effectiveness of \textsc{BadGraph} under adversarial training on the Cora. 
$\Delta$Acc = $\text{Accuracy}_{\text{adv-trained}}-\text{Accuracy}_{\text{vanilla}}$.}
\label{attack_ad}
\begin{tabular}{lccccll} 
\cline{1-5}
Method   & R-GCN                & GCN                  & SGCN                 & TAGCN                &  &   \\ 
\cline{1-5}
Clean    & 86.30                & 85.19                & 87.78                & 87.04                &  &   \\
BadGraph & 72.59                & 75.19                & 75.19                & 75.56                &  &   \\
$\Delta$Acc     & +1.70                & +3.86                & +3.93                & +1.63                &  &   \\ 
\cline{1-5}
         & \multicolumn{1}{l}{} & \multicolumn{1}{l}{} & \multicolumn{1}{l}{} & \multicolumn{1}{l}{} &  &  
\end{tabular}
\end{table}

\subsection{Implementation Details}
We employ GNNs as the graph encoder in all experiments.
Specifically, GAT is used for Cora and Products, while GIN is applied to Arxiv, as defined in Eq.~(1).
For each target node, we retrieve $k=5$ candidate influencer nodes that are semantically distant in the embedding space.
The open-source DeepSeek-V3\footnote{\url{https://github.com/deepseek-ai/DeepSeek-V3}} model serves as the LLM-based attacker, with its API\footnote{\url{https://api.deepseek.com}} generating both topological and textual perturbations.
Throughout this paper, we use the shorthand notations: Qwen for Qwen-Plus, LLaMA for LLaMA-4-17B, and DeepSeek for DeepSeek-V3.
We implement \textsc{BadGraph} in PyTorch\footnote{\url{https://pytorch.org/}}.  
For target backbones we generally use the default hyperparameters from GRB~\cite{zheng2021graph}; when those defaults produced unreasonably low clean accuracies, we made justified adjustments to ensure fair and representative baselines.  
All attack methods—\textsc{BadGraph} and competitors—are evaluated under the same strict black-box setting.  
We implement WTGIA using the authors' official code and adopt its best-performing variant (the TDGIA-perturbed graph after training) \cite{zou2021tdgia}.  
Adversarial texts for injected fake nodes are generated by converting embeddings with LLaMA3-8B \cite{touvron2023llama}.  
The number of injected nodes is 20, 166, and 166 for Cora, OGBN-Products, and OGBN-Arxiv, respectively.  
For consistency across attack settings, all other baselines use a unified configuration: Bag-of-Words (BoW) for node features, GCN as the surrogate model, and up to two edge perturbations per node.  
The perturbed graphs produced by these procedures are then evaluated on the same set of target models.  
Public implementations of the compared methods are available at the referenced repositories.

\paratitle{Target backbones:}
\begin{itemize}[leftmargin=*]
\item R-GCN, GCN, GIN, GraphSAGE, SGCN, TAGCN\footnote{\url{https://github.com/thudm/grb}}.
\item RUNG\footnote{\url{https://github.com/chris-hzc/RUNG}}.
    \item LLM4RGNN\footnote{\url{https://github.com/zhongjian-zhang/LLM4RGNN}}.
  \end{itemize}

\paratitle{Baselines:}
  \begin{itemize}[leftmargin=*]
    \item WTGIA\footnote{\url{https://github.com/Leirunlin/Text-level-Graph-Attack}}.
    \item STACK, PGD, RND, FLIP\footnote{\url{https://github.com/thudm/grb}}.
    \item SGAttack, NETTACK\footnote{\url{https://github.com/DSE-MSU/DeepRobust}}.
  \end{itemize}

\subsection{Computing Environment and Resources}
The experiments are conducted in a computing environment with the following specifications:
\begin{itemize}[leftmargin=*]
    
    \item OS: Ubuntu 22.04.3 LTS (Linux kernel 5.15.0-124-generic)
   \item CPU: Intel(R) Xeon(R) Platinum 8360Y CPU @ 2.40GHz
   \item GPU: NVIDIA GeForce RTX 4090, 24GB
     
\end{itemize}

\section{More  Experimental Results}

\subsection{Evaluating Effectiveness Under Defense}

\textbf{Observation.} \textbf{\textsc{BadGraph} effectively breaks advanced defenses by exploiting cross-modal alignment vulnerabilities.}
We evaluate \textsc{BadGraph} against two strong defenses—RUNG~\cite{hou2024robust} and adversarial training.
As shown in Table~\ref{tab:attack_rung}, \textsc{BadGraph} reduces RUNG’s accuracy by over 50\% under the SBERT encoding, revealing its strong adversarial potency even against robust models.
This degradation stems from semantically aligned cross-modal perturbations that mislead RUNG’s edge reweighting mechanism, causing the model to overemphasize adversarial edges and aggregate corrupted information.
Moreover, as shown in Table~\ref{attack_ad}, \textsc{BadGraph} remains effective under adversarial training, demonstrating strong transferability and resilience to conventional defense strategies.
Together, these results highlight that cross-modal, LLM-guided perturbations can bypass even graph defenses—empirically validating our theoretical claims on cross-modal synergy.

\begin{table}
\centering
\caption{Attack performance with different encoders on the Cora dataset across various GNN encoder backbones.}
\small
\label{tab:different_encoder}
\begin{tabular}{lccccc} 
\hline
Encoder                                                          & GIN            & GraphSAGE      & SGCN           & TAGCN          & DeepSeek        \\ 
\hline
Clean                                                           & 85.70          & 86.00          & 86.07          & 88.22          & 71.85           \\
\begin{tabular}[c]{@{}l@{}}GCN\end{tabular}  & 55.44          & 54.07          & 58.63          & 55.59          & 21.11           \\
\begin{tabular}[c]{@{}l@{}}R-GCN\end{tabular} & \textbf{46.96} & \textbf{38.52} & 57.04          & 54.37          & 19.59           \\
\begin{tabular}[c]{@{}l@{}}GAT\end{tabular}  & 47.04          & 39.93          & \textbf{55.63} & \textbf{51.19} & \textbf{17.03}  \\
\hline
\end{tabular}
\end{table}
\begin{figure}[t]
	{
		\begin{minipage}[t]{0.495\linewidth}
			\centering
			\includegraphics[width=1\textwidth]{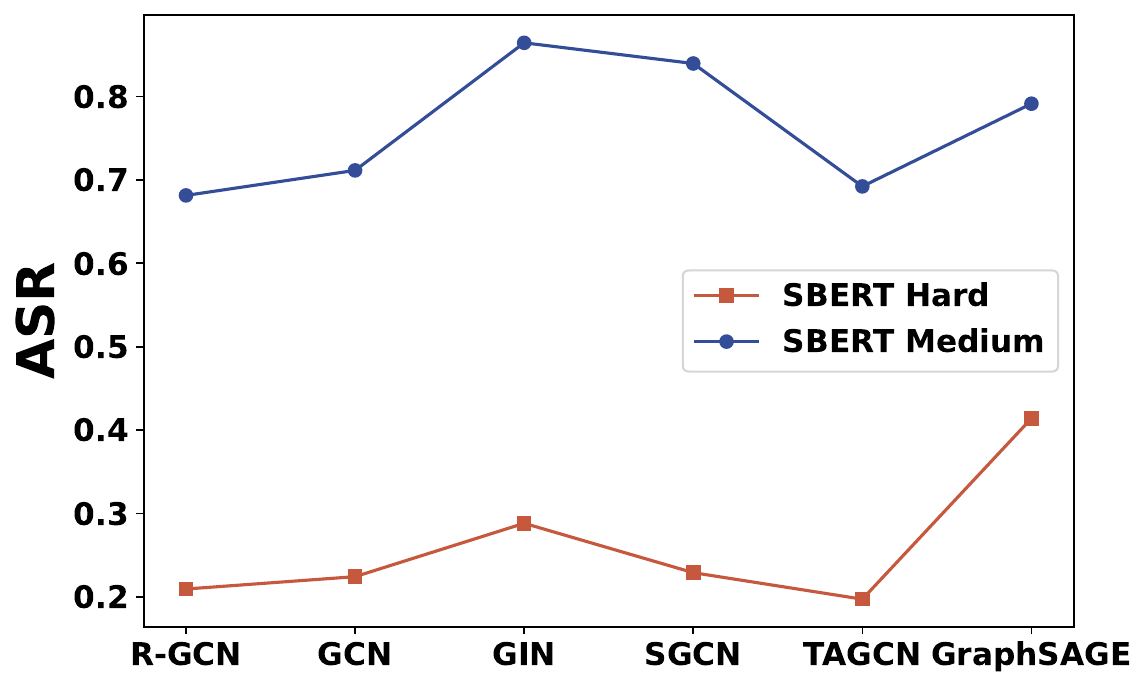}
			% \vspace{-0.1in}
			\subcaption{SBERT}
		\end{minipage}
		\begin{minipage}[t]{0.495\linewidth}
			\centering
			\includegraphics[width=1\textwidth]{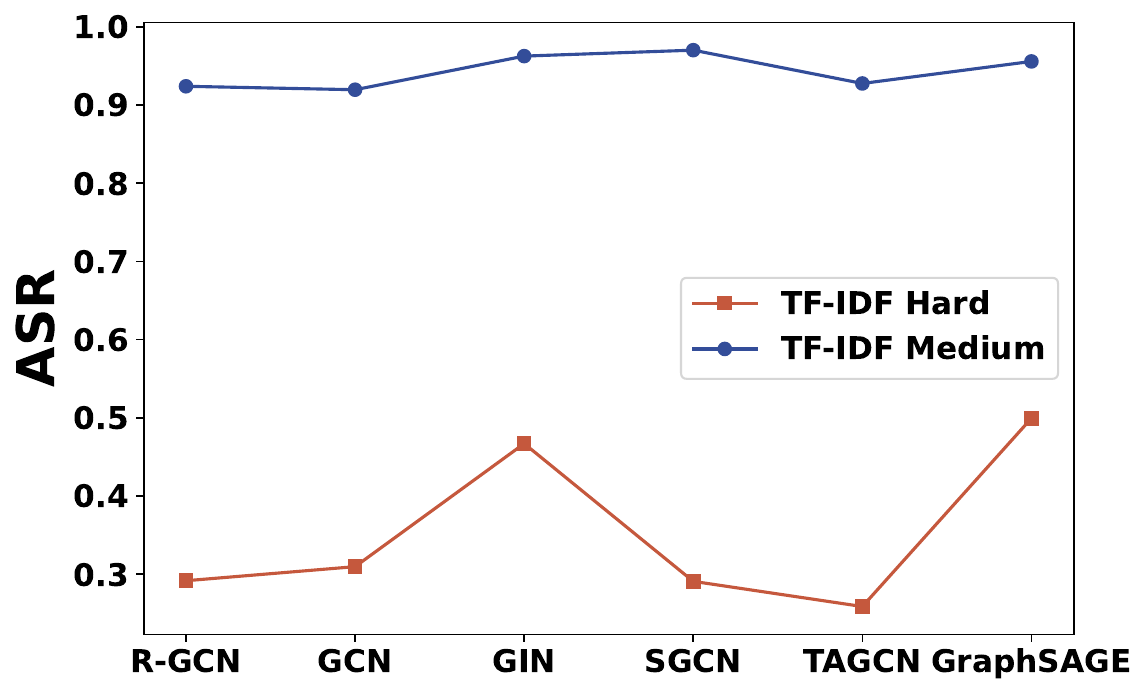}
			% \vspace{-0.1in}
			\subcaption{TF-IDF}
		\end{minipage}
	}
	  \caption{Attack Success Rate on OGBN-Products for different target GNNs, evaluated on two settings: Medium (medium-degree nodes) and Hard (high-degree nodes).}\label{fig:low-degree}
\label{degree}
\end{figure}

\begin{table*}
\centering
\small

\caption{\textbf{GNN-as-Reasoner Scenarios.}  
Comprehensive evaluation results (Average, 3-MAX, and Weighted) across all target backbones.
Lower values indicate stronger attack effectiveness.  
"Clean" denotes unperturbed graphs, while \textit{Our-text} and \textit{Our-struct} represent text-only and structure-only variants, respectively.  
Best results are highlighted in bold.}

\begin{tabular}{cclcccccccccc} 
\hline
\multicolumn{1}{l}{Dataset} & \multicolumn{1}{l}{Emb.} & Metrics & Clean & STACK & PGD & RND & FLIP & SGAttack & WTGIA & Our-text & Our-struct & BadGraph \\ 
\hline
\multirow{9}{*}{\rotatebox{90}{Cora}} 
 & \multirow{3}{*}{\rotatebox{90}{TF-IDF}} 
 & Average  & {\cellcolor[rgb]{0.933,0.933,0.933}}86.19 & 73.14  & 78.51  & 74.07  & 74.19  & 82.38  & 84.95  & 82.60  & 78.90  & \textbf{67.52 } \\
 & & 3-MAX    & {\cellcolor[rgb]{0.933,0.933,0.933}}86.91  & 73.93  & 79.28  & 74.91  & 75.01  & 85.46  & 86.49  & 84.79  & 80.47  & \textbf{72.17 } \\
 & & Weighted & {\cellcolor[rgb]{0.933,0.933,0.933}}87.03  & 74.61  & 79.72  & 74.85  & 75.36  & 86.14 & 86.84  & 84.90  & 81.85  & \textbf{72.51 } \\ 
\hhline{~------------}
 & \multirow{3}{*}{\rotatebox{90}{SBERT}}  
 & Average  & {\cellcolor[rgb]{0.933,0.933,0.933}}86.74   & 76.57  & 80.32  & 76.93   & 76.58  & 83.53 & 85.28   & 77.01   & 77.02   & \textbf{51.07  } \\
 & & 3-MAX    & {\cellcolor[rgb]{0.933,0.933,0.933}}87.56   & 77.75  & 80.67   & 78.07   & 77.26   & 87.14  & 86.72   & 80.35  & 80.15   & \textbf{56.10  } \\
 & & Weighted & {\cellcolor[rgb]{0.933,0.933,0.933}}87.79   & 77.71   & 80.70  & 78.49   & 77.32   & 87.55  & 87.17   & 80.06   & 81.42   & \textbf{55.68  } \\ 
\hhline{~------------}
 & \multirow{3}{*}{\rotatebox{90}{TAPE}}   
 & Average  & {\cellcolor[rgb]{0.933,0.933,0.933}}88.51 & 81.99   & 83.47  & 82.93  & 84.69   & 84.25   & - & 84.60  & 81.09   & \textbf{66.20  } \\
 & & 3-MAX    & {\cellcolor[rgb]{0.933,0.933,0.933}}89.28  & 82.84   & 84.25   & 83.33   & 85.68   & 87.60  & - & 85.73   & 83.23  & \textbf{71.06  } \\
 & & Weighted & {\cellcolor[rgb]{0.933,0.933,0.933}}89.44   & 82.91   & 84.21   & 83.36   & 85.99   & 87.30   & - & 85.91   & 83.39   & \textbf{71.46} \\ 
\hline
\multirow{9}{*}{\rotatebox{90}{Arxiv}} 
 & \multirow{3}{*}{\rotatebox{90}{TF-IDF}} 
 & Average  & {\cellcolor[rgb]{0.933,0.933,0.933}}65.43   & 62.71   & 60.52  & 59.75   & 58.18   & - & 60.38   & 61.56   & 58.07   & \textbf{51.27  } \\
 & & 3-MAX    & {\cellcolor[rgb]{0.933,0.933,0.933}}67.57   & 65.67   & 62.69   & 61.85   & 60.34   & - & 64.24   & 66.13 & 62.13   & \textbf{58.33  } \\
 & & Weighted & {\cellcolor[rgb]{0.933,0.933,0.933}}67.87   & 65.30   & 63.32   & 62.36   & 61.21   & - & 64.09   & 66.30  & 63.53   & \textbf{57.79 } \\ 
\hhline{~------------}
 & \multirow{3}{*}{\rotatebox{90}{SBERT}}  
 & Average  & {\cellcolor[rgb]{0.933,0.933,0.933}}68.56   & 67.68   & 64.75 & 64.81   & 63.43   & - & 65.34   & 66.03  & 62.14   & \textbf{54.60  } \\
 & & 3-MAX    & {\cellcolor[rgb]{0.933,0.933,0.933}}69.78  & 68.27   & 66.27  & 66.12   & 64.59  & - & 67.72   & 67.20   & 65.68  & \textbf{59.51 } \\
 & & Weighted & {\cellcolor[rgb]{0.933,0.933,0.933}}69.91   & 68.19   & 66.46   & 66.60   & 65.01   & - & 68.60   & 67.25   & 66.55   & \textbf{59.25  } \\ 
\hhline{~------------}
 & \multirow{3}{*}{\rotatebox{90}{TAPE}}   
 & Average  & {\cellcolor[rgb]{0.933,0.933,0.933}}74.84   & 77.76 & 73.30  & 74.11  & 73.88   & - & - & 63.58  & 69.72   & \textbf{51.00  } \\
 & & 3-MAX    & {\cellcolor[rgb]{0.933,0.933,0.933}}79.66   & 79.92   & 79.11  & 79.03  & 79.31   & - & - & 69.31  & 76.13   & \textbf{63.57  } \\
 & & Weighted & {\cellcolor[rgb]{0.933,0.933,0.933}}79.59 & 80.51   & 78.60   & 78.74  & 78.88   & - & - & 68.55   & 77.81 & \textbf{63.47  } \\ 
\hline
\multirow{9}{*}{\rotatebox{90}{Products}} 
 & \multirow{3}{*}{\rotatebox{90}{TF-IDF}} 
 & Average  & {\cellcolor[rgb]{0.933,0.933,0.933}}85.66   & 79.60  & 82.95  & 81.83  & 80.79   & 83.44  & 78.67  & 79.21   & 78.63   & \textbf{64.72  } \\
 & & 3-MAX    & {\cellcolor[rgb]{0.933,0.933,0.933}}86.27  & 80.87   & 83.60   & 82.35   & 81.91   & 85.33   & 83.13   & 82.67   & 82.05  & \textbf{71.97  } \\
 & & Weighted & {\cellcolor[rgb]{0.933,0.933,0.933}}86.32   & 81.38   & 83.88   & 82.31   & 82.33   & 85.06 & 83.00   & 82.52   & 81.91   & \textbf{72.13  } \\ 
\hhline{~------------}
 & \multirow{3}{*}{\rotatebox{90}{SBERT}}  
 & Average  & {\cellcolor[rgb]{0.933,0.933,0.933}}87.22  & 82.30  & 85.86  & 84.28   & 83.55   & 85.35 & 81.42   & 85.08   & 81.35   & \textbf{73.97  } \\
 & & 3-MAX    & {\cellcolor[rgb]{0.933,0.933,0.933}}87.64  & 83.13  & 86.32   & 84.65   & 84.19   & 87.13   & 85.20   & 86.49  & 84.65   & \textbf{78.97  } \\
 & & Weighted & {\cellcolor[rgb]{0.933,0.933,0.933}}87.83   & 83.31   & 86.32  & 84.70   & 84.13  & 87.09   & 85.38   & 86.43   & 84.51   & \textbf{79.09  } \\ 
\hhline{~------------}
 & \multirow{3}{*}{\rotatebox{90}{TAPE}}   
 & Average  & {\cellcolor[rgb]{0.933,0.933,0.933}}89.31  & 89.52   & 88.85   & 88.59  & 89.45  & 88.81   & - & 84.75  & 87.86   & \textbf{75.27  } \\
 & & 3-MAX    & {\cellcolor[rgb]{0.933,0.933,0.933}}89.73   & 90.03   & 89.43   & 89.23  & 90.05   & 89.62  & - & 87.12   & 88.92   & \textbf{80.16  } \\
 & & Weighted & {\cellcolor[rgb]{0.933,0.933,0.933}}89.94  & 90.33   & 89.72   & 89.48   & 90.20  & 89.80   & - & 86.99   & 89.14   & \textbf{79.62 } \\
\hline
\end{tabular}
\label{more-exp}
\end{table*}

\subsection{Influence of Graph Encoder Selection}

\textbf{Observation 1.} \textbf{\textsc{BadGraph} exhibits strong and consistent attack performance across diverse graph encoders.}
We assess the impact of encoder choice by substituting the influencer retriever with different GNNs—GCN, R-GCN, and GAT.
As shown in Table~\ref{tab:different_encoder}, \textsc{BadGraph} consistently causes over 27\% accuracy degradation across all variants, demonstrating its robustness and flexibility.
This adaptability shows that \textsc{BadGraph} can seamlessly integrate with any expressive encoder, making it a general plug-and-play attack framework for diverse graph learning backbones.
\textbf{Observation 2.} \textbf{Encoder expressiveness amplifies cross-model transferability.}
More expressive encoders (e.g., GAT) produce finer-grained relational embeddings, enabling more accurate influencer retrieval and directionally aligned perturbations.
Consequently, \textsc{BadGraph} achieves stronger and more transferable attacks across both GNN- and LLM-based reasoners.
% Even when attacking LLMs, all encoders remain effective, but GAT yields the most substantial degradation by guiding the LLM toward richer, semantically coherent perturbations.
These findings confirm that encoder expressiveness enhances both the performance and generalization of LLM-driven adversarial attacks, reinforcing \textsc{BadGraph}’s universality across architectures and modalities.

\subsection{Low-Degree Nodes Are More Vulnerable to Adversarial Perturbations}

\textbf{Observation.} \textbf{Nodes with fewer neighbors exhibit significantly higher attack susceptibility.}
We analyze the relationship between node degree and adversarial vulnerability on the Products dataset under the GNN-as-Reasoner setting, comparing \emph{medium} (low-degree) and \emph{hard} (high-degree) splits.
As shown in Fig.~\ref{degree}, the Attack Success Rate (ASR) is markedly higher for the medium split, indicating that low-degree nodes are more easily compromised by adversarial perturbations.
These findings highlight the necessity for defense mechanisms that explicitly account for the heightened vulnerability of low-degree nodes in TAG security scenarios.

\subsection{Additional Experimental Results.}
Due to space limitations in the main paper, we omitted part of Table~\ref{tab:overall-performance}.  
Specifically, the detailed results for the three evaluation metrics—Average, 3-MAX, and Weighted—were excluded.  
Table~\ref{more-exp} in this appendix provides the complete results across all datasets and evaluation metrics.  
Overall, \textsc{BadGraph} consistently achieves state-of-the-art performance, demonstrating strong effectiveness and generalizability under diverse settings.

\end{document}